\documentclass[preprint,11pt]{elsarticle}

\usepackage[a4paper,
            bindingoffset=0.2in,
            left=1.155in,
            right=1.155in,
            top=1.185in,
            bottom=1.185in,
            footskip=.25in]{geometry}

\usepackage{amssymb}
\usepackage{amsfonts}
\usepackage{nicefrac}

\usepackage{times}
\usepackage{float}
\usepackage{amsmath}
\usepackage{changepage}
\usepackage{graphicx}
\usepackage{caption}
\usepackage{relsize}
\usepackage{times}
\usepackage{natbib}
\usepackage{hyperref}
\usepackage{mathtools}

\usepackage{multirow}

\DeclarePairedDelimiter\abs{\lvert}{\rvert}%
\makeatletter
\let\oldabs\abs
\def\abs{\@ifstar{\oldabs}{\oldabs*}}
\let\oldnorm\norm
\def\norm{\@ifstar{\oldnorm}{\oldnorm*}}
\makeatother

\journal{Automation in Construction}

\makeatletter
\def\ps@pprintTitle{%
 \let\@oddhead\@empty
 \let\@evenhead\@empty
 \def\@oddfoot{}%
 \let\@evenfoot\@oddfoot}
\makeatother

\usepackage{array}
\newcolumntype{L}{>{\centering\arraybackslash}m{1.35cm}}

\begin{document}

\begin{frontmatter}



\title{Safer Together: Machine Learning Models Trained on Shared Accident Datasets Predict Construction Injuries Better than Company-Specific Models \\ {\small\textit{Submitted to Automation in Construction}}}

\author[label1]{Antoine J.-P. Tixier\footnote{antoine.tixier@safetyfunction.com}}
\author[label1,label2]{Matthew R. Hallowell}
\address[label1]{SafetyAI R\&D}
\address[label2]{University of Colorado at Boulder}

\begin{abstract}

\small

\textbf{Highlights}

\begin{itemize}
    \item 9 companies from 3 domains (construction, electric T\&D, oil \& gas) shared their accident datasets.
    \item Machine learning models were trained to predict safety outcomes from fundamental attributes.
    \item Models trained on all datasets (full generic models) outperformed the company-specific models in 82\% of the company-domain-outcome combinations, with large gains in F1 score (+4.4 on average and up to +15.3).
    \item On average, generic models predicted 2.26 categories more than specific models (up to 7), making for more useful forecasts in practice.
    \item Per-domain generic models were not always better than full generic models.
    \item Combining generic and specific models (data quantity and relevance) was often very beneficial.
    \item Generic models give companies devoid of accident datasets access to safety predictions.
    \item Generic models address safety cross-organizational learning and dissemination in construction.
\end{itemize}

In this study, we capitalized on a collective dataset repository of 57k accidents from 9 companies belonging to 3 domains and tested whether models trained on multiple datasets (generic models) predicted safety outcomes better than the company-specific models.
We experimented with full generic models (trained on all data), per-domain generic models (construction, electric T\&D, oil \& gas), and with ensembles of generic and specific models.
Results are very positive, with generic models outperforming the company-specific models in most cases while also generating finer-grained, hence more useful, forecasts.
Successful generic models remove the needs for training company-specific models, saving a lot of time and resources, and give small companies, whose accident datasets are too limited to train their own models, access to safety outcome predictions.
It may still however be advantageous to train specific models to get an extra boost in performance through ensembling with the generic models.
Overall, by learning lessons from a pool of datasets whose accumulated experience far exceeds that of any single company, and making these lessons easily accessible in the form of simple forecasts, generic models tackle the holy grail of safety cross-organizational learning and dissemination in the construction industry.

\end{abstract}

\begin{keyword}
construction safety \sep artificial intelligence \sep supervised learning \sep injury prediction \sep transfer learning \sep data sharing \sep collective intelligence


\end{keyword}

\end{frontmatter}


\section{Introduction}
The SafetyAI council is a community of large organizations from the construction, oil \& gas, and electric Transmission and Delivery (T\&D) domains, that share their safety-related data with the SafetyAI Research and Development (R\&D) team.

Before exploiting the data, the R\&D team is in charge of standardizing the datasets received by each company, which is crucial, as each one features different variables and different category names for each variable.
Standardization makes sure that all datasets are based on the same taxonomy, i.e., speak the same language.

The SafetyAI community dataset, comprising close to a million events including near misses, observations, good catches, etc., is only accessible to the R\&D team, a neutral party, which guarantees that it is impossible for companies to see each other's data, and that the output of all the R\&D conducted on the collective dataset is made available to the entire community.
This is of paramount importance, in a very competitive environment.

In this study, we started by extracting attributes from accident reports.
We briefly introduce the attribute framework in what follows.

\subsection{Attribute-based framework}
Attributes are basic descriptors of construction work that are observable \textit{before} accident occurrence, and cover means, methods, and environmental conditions \cite{desvignes2014requisite,villanova2014attribute}.
One advantage of the attribute-based framework over modeling at the task or work package level is that attributes are fundamental and universal. That is, any situation from any site around the world, in any industry sector, can be characterized by a set of attributes.
Attributes can be recorded on-the-fly on site, or can be extracted offline from various mediums such as photos and text reports.
For instance, four attributes can be extracted from the narrative \texttt{\small worker tripped on a cable when carrying a 2x4 to his truck}: (1) cable, (2) object on the floor, (3) lumber, and (4) light vehicle.

Narratives are particularly well-suited if the goal is to use attributes for predictive modeling.
Indeed, in incident report databases, narratives are often paired with outcomes such as accident type, injury severity, body part impacted, etc.
Attributes also completely anonymize narratives, which is especially desirable when considering a pool of datasets aggregated from different companies.
For any given event, everything that remains is a set of attributes and a set of standardized safety outcomes.

However, manually extracting attributes from large amounts of text reports is very costly in terms of human resources and pose inter-annotator agreement issues.
To solve this problem, we developed and validated a Natural Language Processing (NLP) tool based on rules and lexicons \cite{tixier2016automated}.
We later proved that using the attributes extracted by the tool to predict safety outcomes was effective and valid \cite{tixier2016application,baker2020ai}.
We also used the attributes extracted by the tool for unsupervised learning applications, such as clustering and visualization \cite{tixier2017clash}, and risk modeling and simulation \cite{tixier2017construction}.

\subsection{Differences with our previous research and objective of the current study}
In our original study \cite{tixier2016application}, we provided a proof for the concept of predicting safety outcomes from attributes, both extracted with the NLP tool.
Then, in \cite{baker2020ai}, we showed that attributes were still highly predictive when the safety outcomes were given by independent human annotations, which definitely validated the approach.
We also used a much larger dataset than in the original study, two new supervised learning algorithms, model stacking, a healthier experimental setup with more appropriate performance metrics, and we analyzed per-category attribute importance scores.
We also showed that unlike what we had concluded in \cite{tixier2016application}, injury severity was predictable from attributes.

In the present research, we interested ourselves with a new, completely different problem.
We had access to a pool of accident datasets coming from 9 companies, and our goal was to: \\

\noindent\fbox{%
\parbox{\textwidth}{%
\textit{``Test whether predictive models trained on a \textbf{generic} dataset (i.e., aggregated from the datasets of multiple companies) outperformed the models trained on the \textbf{specific} dataset of each company.''}
}%
}

\vspace{0.5cm}

More precisely, we experimented with two types of generic models:

\begin{itemize}
\item \textbf{Full} generic model: one model trained on the datasets of all companies.
\item \textbf{Per-domain} generic models: one model per industry sector, trained only on the datasets of the companies involved in that sector (or the parts thereof, as some companies belong to multiple domains).
\end{itemize}

The potential advantages of generic models are numerous:

\begin{enumerate}
\itemsep1em 
\item Usually with machine learning, the more data, the better, so generic models are expected to bring improvements in predictive skill compared to the company-specific models.
This is not guaranteed however, as one important question is whether (1) more data (generic datasets) or (2) more relevant data (specific datasets) is better.
\item By being trained on larger datasets, the generic models learn to predict a greater variety of outcome categories than the specific models, making for more useful forecasts.
\item Successful generic models would remove the needs for training specific models for each company, saving a lot of time and resources.
\item Alternatively, if company-specific models are already available, combining them with the generic models may provide an extra boost in performance.
\item Last but not least, successful generic models would give small companies -whose accident datasets are too limited to train their own specific models- access to high quality safety outcome forecasts.
\end{enumerate}

From a high level, generic models tackle the holy grail of safety cross-organizational learning and dissemination in the construction industry.
Indeed, generic models (1) learn lessons from a pool of datasets whose quantity and diversity\footnote{Diversity of situations, means and methods, environmental conditions, geographical areas...} of accumulated experience far exceeds that of any single company, and (2) disseminate these lessons as forecasts, which are clear, direct, and easily accessible information, via, e.g., a user interface (desktop or mobile) or API taking attributes as input and returning probabilities for each category of each outcome.

Moreover, one should note that in the pool, the individual biases of each dataset, due to specific annotators, reporting practices and policies, etc., tend to average out.
Consequently, the lessons learned by the supervised learning algorithms on the generic datasets are more objective and broadly applicable than that learned on the specific datasets.

\section{Background}
The needs to share standardized incident data at the industry level to enable collaborative learning have long been recognized in aviation and transportation \cite{tanguy2016natural}.
Some examples include the NASA-managed Aviation Safety Reporting System (ASRS) database, created in 1976 and featuring over a million incidents, or the European Coordination Center for Accident and Incident Reporting Systems (ECCAIRS) database, started in 2004.
Such collective repositories also exist in the chemical industry, with the Major Accident Reporting System (eMARS) of the European Commission, launched in 1982, and the Process Safety Incident Database (PSID) of the Center for Chemical Process Safety \cite{sepeda2006lessons}.\\

However, the construction industry still lacks comparable initiatives.
The needs for data storage and access infrastructures for construction safety did start to receive some attention recently \cite{le2014social,pedro2022data}, but most efforts placed themselves at the company or project level.
Cross-organizational safety data collection is still rare in practice \cite{edwin2022sharing,wasilkiewicz2018information}.
This is a major issue, as collaborative machine learning at the industry level is not possible until a common pool of standardized datasets has been put together.
This provided the motivation for us to create the SafetyAI council in 2020.\\

One should note that some consortiums already exist, such as the INGAA Foundation, the Edison Electric Institute (EEI), the Construction Safety Research Alliance (CSRA), or the National Safety Council (NSC), but their activities do not revolve around systematic large-scale accident data collection and analysis.
These initiatives rather involve working groups, communities of practice, qualitative analyses, and conferences, towards building communications, policies, best practices, business intelligence, safety culture and leadership, training material, etc.
In other words, they are based on ``soft'' methods for knowledge sharing and collaborative learning at the human level.
They do not primarily conduct ``hard'' scientific research and software development, and do not pool accident datasets for AI applications and automatic large-scale learning and dissemination.

\section{Data Description}
As already explained, as part of the SafetyAI initiative, we had access to a pool of safety datasets coming from nine large companies from the construction, oil \& gas, and electric Transmission and Delivery (T\&D) domains.
One company, Company7\footnote{Company names have been anonymized.}, also had about 600 corporate services (office) events for the severity outcome.
We kept these cases as training data for the full generic model but did not train a specific model on them.

Member companies conduct work mostly in North America, and rely on their own teams as well as contractors.
The collective dataset covers the period 2000 to 2022, with a distribution biased towards the last decade and especially more recent years.

While the entire pool comprises almost a million events including near misses and observations, we focused on accident cases only in this effort.
As can be seen in Table \ref{table:overview}, the sizes of the individual datasets ranged from 2k to 20k cases, with an average of 6k per company.
There were 57262 accident cases in total, recorded over tens of millions of work hours.

\begin{table}[ht]
\centering
\scalebox{0.8}{
\begin{tabular}{r|LLLLLLLLL}
\hline
 & Comp.1 & Comp.2 & Comp.3 & Comp.4 & Comp.5 & Comp.6 & Comp.7 & Comp.8 & Comp.9 \\
\hline
Domains & Constr., elec. & Oilgas & Constr., oilgas & Elec. & Constr. & Constr., elec. & Elec., oilgas, corp. & Oilgas & Elec. \\
\hline
Regions & Canada & California & NAM & NAM & NAM & NAM & NAM, Mexico & World$^\star$ & Southeast USA \\
\hline
$n$ & 4481 & 1965 & 4072 & 5321 & 7245 & 4310 & 8345 & 19298 & 2225 \\
\hline
\end{tabular}
}
\caption{Company overview.
NAM: North America (Canada + USA).
Constr.: construction.
Elec: electric T\&D.
Oilgas: oil \& gas.
$^\star$Including ships and rigs.
Corp: corporate.}
\label{table:overview}
\end{table}

We considered the same outcomes as in \cite{baker2020ai}: injury severity, body part impacted, injury type, and accident type.
The columns corresponding to each outcome were selected from the company datasets and normalized to use a common, standard set of categories, shown in Table \ref{table:categories}.
Not all outcomes were available for every event of every company.
From the narrative of each report, we extracted with the NLP tool \cite{tixier2016automated} the original set of 80 attributes \cite{tixier2016automated,baker2020ai}, plus 11 new items (see Table \ref{table:atts}).
We also used the tool to extract a fifth outcome, energy source, that was not available in the company datasets.

\begin{table}[H]
\centering
\scalebox{0.75}{
\begin{tabular}{llllllllll}
  \hline
\multicolumn{2}{c}{Injury Severity} & \multicolumn{2}{c}{Body Part} & \multicolumn{2}{c}{Injury Type} & \multicolumn{2}{c}{Accident Type} & \multicolumn{2}{c}{Energy Source} \\ 
  \hline
first aid & 38994 & hand & 15782 & cut & 14086 & handling & 6379 & motion & 33958 \\ 
  report-only & 6993 & head & 10296 & strain & 10069 & fall & 5374 & gravity & 15904 \\ 
  lost time & 5319 & leg & 6550 & contusion & 8558 & exposure & 3986 & chemical & 2411 \\ 
  medical & 4913 & arm & 5943 & foreign body & 3348 & struck & 3834 & biological & 2044 \\ 
  recordable & 1043 & trunk & 5375 & pinch & 1756 & contact & 2269 & thermal & 1691 \\ 
   &  & foot & 4632 & fracture & 1681 & caught & 1758 & mechanical & 611 \\ 
   &  & multiple/entire & 942 & burn & 1454 & overexertion & 1523 & pressure & 296 \\ 
   &  &  &  & irritation & 1222 & equipment & 1449 & electricity & 181 \\ 
   &  &  &  & pain & 1194 & PPE & 949 & radiation & 166 \\ 
   &  &  &  & exhaustion & 1054 & transitioning & 578 &  &  \\ 
   &  &  &  & bite & 710 & error & 425 &  &  \\ 
   \hline
\end{tabular}
}
\caption{Outcome category counts, across all companies and domains.
PPE: personal protective equipment.}
\label{table:categories}
\end{table}

\section{Experimental Setup}

\subsection{Splits}
Train, validation and test splits were created for each of the 51 company-domain-outcome combinations for which at least 2 categories with more than 100 observations each were available (shown in Table \ref{table:data_details}), by randomly sampling without replacement 64\%, 16\%, and 20\% of cases, respectively.
The counts summed over companies are shown in Table \ref{table:splits_summary}.
Note that the proportions we used in our previous work \cite{baker2020ai} were 81\%, 9\% and 10\%, but in the present research, we decided to reserve more observations for the validation and test sets to make them more representative of the training sets, in order to increase the stability and validity of hyperparameter tuning and evaluation\footnote{Increasing the sizes of the validation and test sets was a good alternative to $k$-fold cross-validation, which would have taken too much time.}.

\begin{table}[h]
\centering
\small
\scalebox{0.865}{
\begin{tabular}{rccccc}
\hline
& & \# Companies & Train & Val & Test  \\ 
\hline
\multirow{5}{*}{\rotatebox[origin=c]{90}{{Severity}}}
& Construction & 4 & 9980 & 2494 & 3119 \\ 
& Electric T\&D & 4 & 6672 & 1669 & 2085 \\
& Oil \& Gas & 4 & 18381 & 4595 & 5744 \\
& Corporate & 1 & 418 & 105 & 131 \\
& Full & 9 & 35451 & 8863 & 11079 \\ 
\hline
\multirow{4}{*}{\rotatebox[origin=c]{90}{{\footnotesize Body Part}}}
& Construction & 4 & 8209 & 2052 & 2565 \\ 
& Electric T\&D & 4 & 6036 & 1508 & 1885 \\ 
& Oil \& Gas & 3 & 15788 & 3947 & 4933 \\ 
& Full & 9 & 30033 & 7507 & 9383 \\ 
\hline
\multirow{4}{*}{\rotatebox[origin=c]{90}{{\footnotesize Injury Type}}}
& Construction & 4 & 6267 & 1566 & 1958 \\ 
& Electric T\&D & 4 & 4764 & 1191 & 1489 \\ 
& Oil \& Gas & 3 & 14960 & 3740 & 4675 \\
& Full & 9 & 25991 & 6497 & 8122 \\ 
\hline
\multirow{4}{*}{\rotatebox[origin=c]{90}{{\scriptsize Acc. Type}}}
& Construction & 2 & 2740 & 685 & 856 \\ 
& Electric T\&D & 2 & 1600 & 400 & 500 \\ 
& Oil \& Gas & 3 & 2910 & 728 & 910 \\
& Full & 6 & 7250 & 1813 & 2266 \\ 
\hline
\multirow{4}{*}{\rotatebox[origin=c]{90}{{\scriptsize En. Source}}}
& Construction & 4 & 4875 & 1218 & 1524 \\ 
& Electric T\&D & 3 & 2637 & 660 & 825 \\ 
& Oil \& Gas & 2 & 2600 & 650 & 813 \\ 
& Full & 8 & 10112 & 2528 & 3162 \\ 
\hline
\end{tabular}
}
\caption{Split counts for each domain-outcome combination, summed over companies.
For \# Companies, full $\neq$ total as some companies belong to multiple domains (see Tables \ref{table:overview} and \ref{table:data_details}).}\label{table:splits_summary}
\end{table}

A specific model was trained on each of the 51 company-domain-outcome combinations for which sufficient data were available, except for that one combination involving the corporate cases, making for a total of 50 specific models.

\begin{table}[h]
\centering
\small
\scalebox{0.81}{
\begin{tabular}{r|ccccc|ccccc|ccccc|c}
\hline
&
\multicolumn{5}{c}{Construction} &
\multicolumn{5}{c}{Electric T\&D} & 
\multicolumn{5}{c}{Oil \& Gas} &
\multicolumn{1}{c}{Corp.} \\
\hline
Comp. & S & B & IT & AT & E & S & B & IT & AT & E & S & B & IT & AT & E & S \\ 
\hline
1 & x & x & x & & x & & & & & & & & & & & \\
2 & & & & & & & & & & & x & x & x & & \\
3 & x & x & x & x & x & & & & & & x & & & x & & \\
4 & & & & & & x & x & x & x & x & & & & & & \\
5 & x & x & x & x & x & & & & & & & & & & & \\
6 & x & x & x & & x & x & x & x & & x & & & & & & \\
7 & & & & & & x & x & x & & & x & x & x & x & x & x \\
8 & & & & & & & & & & & x & x & x & x & x & \\
9 & & & & & & x & x & x & x & x & & & & & & \\
\hline
\end{tabular}
}
\caption{The 51 company-domain-outcome combinations associated with at least 2 categories with more than 100 observations each.
S: severity, B: body part, IT: injury type, AT: accident type, E: energy source. Corp.: corportate.}\label{table:data_details}
\end{table}

For a given domain and a given outcome, the splits of the per-domain generic model were obtained by combining, across all companies, the splits corresponding to that domain and that outcome.
In total, there was one per-domain generic model for each domain and for each outcome, hence a total of $3 \times 5 = 15$ per-domain generic models.

For a given outcome, the splits of the full generic model were obtained by combining, across all companies and across all domains, the splits corresponding to that outcome.
In total, there was one full generic model for each outcome, hence a total of 5 full generic models.

For each of the aforementioned cases, we tried 3 different algorithms, as will be explained in subsection \ref{subsec:ml}.
Hence, a total of $(15 + 5) \times 3 = 60$ generic models were trained.

\subsection{Class imbalance}
To address the problem of class imbalance, weights inversely proportional to category counts in the training set were computed with the formula \texttt{\small max(counts)/counts}, like in \cite{baker2020ai}.
During training, these weights forced the models to pay more attention to the cases from the minority categories.
Per-category counts with training weights can be found in Tables \ref{table:splits} and \ref{table:splits_1} for the 15 domain-outcome combinations.

\subsection{Supervised learning algorithms}\label{subsec:ml}
Like in \cite{baker2020ai}, we relied on three popular machine learning models: Random Forest (RF) \cite{breiman2001random}, eXtreme Gradient Boosting (XGBoost or XGB) \cite{chen2016xgboost}, and linear Support Vector Machine (SVM) \cite{boser1992training}.
More precisely, we used the Python's \texttt{scikit-learn} implementations of Random Forest\footnote{\href{https://scikit-learn.org/stable/modules/generated/sklearn.ensemble.RandomForestClassifier.html}{https://scikit-learn.org/stable/modules/generated/sklearn.ensemble.RandomForestClassifier.html}} and linear SVM\footnote{\href{https://scikit-learn.org/stable/modules/generated/sklearn.svm.LinearSVC.html}{https://scikit-learn.org/stable/modules/generated/sklearn.svm.LinearSVC.html}}, while, for XGBoost, we used the original Python library\footnote{\href{https://xgboost.readthedocs.io/en/latest/python/python_api.html\#module-xgboost.sklearn}{https://xgboost.readthedocs.io/en/latest/python/python\_api.html\#module-xgboost.sklearn}} and in particular the GPU-accelerated implementation of the ``fast histogram'' algorithm (\texttt{gpu\_hist}) as the tree method\footnote{\href{https://xgboost.readthedocs.io/en/latest/gpu/}{https://xgboost.readthedocs.io/en/latest/gpu/}}.

For theoretical details about each algorithm, we refer the reader to our paper \cite{baker2020ai}, publicly available\footnote{\url{https://arxiv.org/pdf/1908.05972.pdf}}.

\subsection{Hyperparameter optimization}
We tuned the models by performing grid searches on the validation sets.
Details about the parameters searched are available in \ref{sec:hyper}.
The final models were trained on the union of the training and validation sets with the best parameter values.
Both the specific models and the generic models were tested on the test sets of the specific models, to ensure fair comparison.
As already explained, there were 50 such test sets, one for each company-domain-outcome combination. 

\subsection{Transfer learning by stacking generic and specific models}\label{sec:stack}
As was mentioned in the introduction, one important question is the extent to which (1) more data (generic datasets) or (2) more relevant data (specific datasets) is better.
In what follows, we explore a way to move past this binary choice and have a tradeoff between quantity and relevance.

Inspired by transfer learning, which is very successful in computer vision \cite{krizhevsky2012imagenet} and NLP \cite{radford2018improving,devlin2018bert,lewis2019bart,eddine2020barthez}, we experimented with combining the predictions of the generic and specific models via an ensemble model.

Very briefly, in AI, transfer learning refers to a two-step process.
First, a model is trained at solving a general task on large amounts of data.
This phase is called the pretraining phase, as it allows the model to acquire generic knowledge (e.g., in NLP, reading and writing), that is applicable to a great variety of situations downstream.
Second, the pretrained model is finetuned on a specific task of interest, often associated with a much smaller dataset (e.g. in NLP, summarization, classification, question answering, paraphrase detection, etc.).

In our case, the generic and the specific models have to perform the same task, i.e., predicting a given safety outcome\footnote{The generic model has to perform a more difficult version of the task, though (more categories to predict).}, and there is no pretraining phase \textit{per se}, in that the generic and the specific models are two different models.
However, our approach is similar in spirit to transfer learning, as our goal is to capitalize on generic knowledge gained from large amounts of data to improve performance on a specific task associated with a smaller dataset.\\

More precisely, for each company-domain-outcome combination, we trained a meta-model taking as input the weighted elementwise sum of the probabilistic forecasts of the best generic and specific models\footnote{The entries of the specific model vector for the categories that it did not predict were set to zero.}.
We used a simple logistic regression\footnote{\href{https://scikit-learn.org/stable/modules/generated/sklearn.linear_model.LogisticRegression.html}{https://scikit-learn.org/stable/modules/generated/sklearn.linear\_model.LogisticRegression.html}} as our meta-model, with the $C$ parameter fixed and equal to $0.2$, like in \cite{baker2020ai}.
We grid searched the validation set to find the best values of coefficients $a$ and $b$ where:
\begin{equation}
\text{input}_\text{ensemble} = a \times \text{output}_\text{generic} + b \times \text{output}_\text{specific} 
\end{equation}

Besides performance considerations, using tunable weights improves interpretability, by providing information regarding which of the generic model or the specific model makes the most important contribution to predictive skill.

We tried values from $0.1$ to $1$ with $0.1$ steps, holding the other parameter equal to $1$, and conversely.
That is, the following 19 pairs: $(0.1,1)$, $(0.2,1)$, ... , $(1,1)$, $(1,0.1)$, $(1,0.2)$, ... , $(1,0.9)$. \\

\textbf{SVM issue}.
By design, the implementation of the linear SVM model we used, \texttt{\small linearSVC}, only returns discrete predictions, that is, a single label corresponding to the most likely category, rather than a probability distribution over all categories.
To address this issue, in \cite{baker2020ai}, we tried retraining the best SVM using the \texttt{\small SVC} implementation\footnote{ \href{https://scikit-learn.org/stable/modules/generated/sklearn.svm.SVC.html}{https://scikit-learn.org/stable/modules/generated/sklearn.svm.SVC.html}} with linear Kernel.
However, results were not convincing.
Therefore, in the present study, we decided simply not to use model stacking when one of the two models involved (e.g., best generic or specific model) was a SVM.

\subsection{Performance metrics}
Due to the large class imbalance for all outcomes, measuring classification performance with accuracy was inadequate. Rather, we computed precision, recall, and F1-score.

Precision, respectively recall, for category $i$, is equal to the number of correct predictions for category $i$ (number of hits), divided by the number of predictions made for category $i$ (hits and false alarms), respectively by the number of observations in category $i$ (hits and misses).

\begin{equation}
\textrm{precision} = \frac{C_{i,i}}{\sum_{j=1}^{K}C_{j,i}} \hspace{2cm}
\textrm{recall} = \frac{C_{i,i}}{\sum_{j=1}^{K}C_{i,j}}
\end{equation}

\vspace{0.235cm}

Where the confusion matrix $C$ is a square matrix of dimension $K \times K$ ($K$ being the number of categories) and whose $(i,j)^{th}$ element $C_{i,j}$ indicates how many of the observations known to be in category $i$ were predicted to be in category $j$. 
Finally, we computed the F1-score, the harmonic mean of precision and recall:
\begin{equation}
\text{F1} = 2 \times \frac{\text{precision} \times \text{recall}}{\text{precision} + \text{recall}}
\end{equation}

\subsection{Configuration}
We relied on a single Ubuntu 20.04.4 machine featuring a 4.9 MHz 12-thread CPU, a 12 GB Nvidia Titan V GPU, 64 GB of RAM, R version 4.1.3 \cite{r_language}, and Python version 3.8.13 with \texttt{\small scikit-learn} version 1.1.1 \cite{scikit-learn}.
Running all experiments took approximately ten days.

\section{Results}
Each generic model (full and per-domain), as well as ensembles thereof (stacking approach described in section \ref{sec:stack}) was tested on the test set of each company-domain-outcome combination and compared against the best performing specific model for this combination.

Results are very positive.
As can be seen in Table \ref{table:data_gains_comp}, across all companies, the generic models (full or per-domain) outperform the specific models 82\% of the time, i.e., for 41 company-domain-outcome combinations out of 50.
Detailed per-company results can be found in \ref{sec:full_details} for the full generic models and \ref{sec:generic_details} for the per-domain generic models.
At the company level, improvements are brought on average for 80.6\% of outcomes (across all domains), ranging from 33.3\% for Company2 to 100\% for Company1, Company4, Company6, and Company9.

\begin{table}[H]
\centering
\large
\scalebox{0.6735}{
\begin{tabular}{r|ccccc|ccccc|ccccc}
\hline
&
\multicolumn{5}{c}{Construction} &
\multicolumn{5}{c}{Electric T\&D} & 
\multicolumn{5}{c}{Oil \& Gas} \\
\hline
C & S & B & IT & AT & E & S & B & IT & AT & E & S & B & IT & AT & E \\ 
\hline
1 & +1.26 & +0.2 & +2.55 & & +3.07 & & & & & & & & & & \\
2 & & & & & & & & & & & x & +3.15 & x & \\
3 & x & +6.56 & +0.49 & x & +12.47 & & & & & & +0.99 & & & x & \\
4 & & & & & & +3.49 & +0.47 & +3.06 & +0.98 & +0.63 & & & & & \\
5 & x & +2.64 & +1.29 & +3.14 & +2.79 & & & & & & & & & & \\
6 & +12.86 & +4.39 & +1.63 & & +7.09 & +12.87 & +5 & +11.19 & & +0.59 & & & & & \\
7 & & & & & & x & +6.69 & +15.3 & & & x & +1.04 & +9.54 & +2.12 & +2.2 \\
8 & & & & & & & & & & & +1.11 & +0.47 & +2.78 & x & +3.13 \\
9 & & & & & & +1.56 & +5.38 & +12.16 & +5.01 & +6.16 & & & & & \\
\hline
\end{tabular}
}
\caption{Company-level max gains. x: no improvement.
S: severity, B: body part, IT: injury type, AT: accident type, E: energy source. C: company
}\label{table:data_gains_comp}
\end{table}

\begin{figure}[H]
\centering
\includegraphics[width=0.35\textwidth]{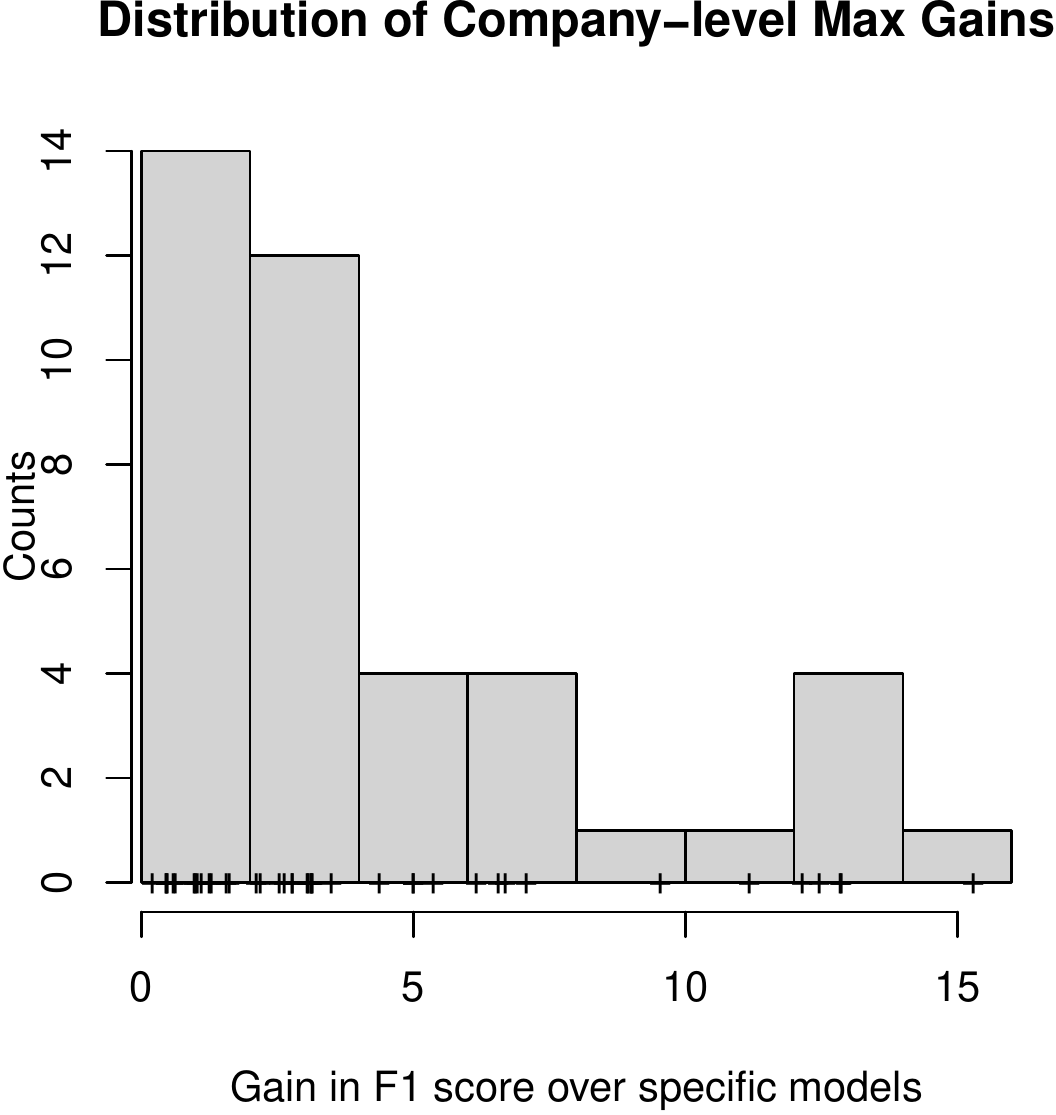}
\captionsetup{size=footnotesize}
\caption{Company-level max gains, across all domains and outcomes. n=41, min=0.2, max=15.3, mean=4.4. \label{fig:company_gains}}
\end{figure}

Furthermore, as shown in Fig. \ref{fig:company_gains}, gains are high on average (+4.4 in F1 score) and reach impressive values, e.g., +15.3 for Company7 on electric T\&D-injury type, +12.87 for Company6 on electric T\&D-severity, +12.86 for Company6 on construction-severity, +6.56 for Company3 on construction-body part, etc.
And all of that, while predicting more categories.

There are only 9 domain-outcome combinations over 50, across 5 companies, on which the generic models do not bring any quantitative improvement.
However, since their forecasts are more informative (more categories predicted), it may still make sense in practice to use the generic models in lieu of the specific models, even on these combinations.
For instance, for Company3-oil \& gas-accident type, the specific model only predicts \textit{exposure} and \textit{struck}, but the generic model also predicts the categories \textit{caught}, \textit{fall}, and \textit{overexertion}.\\

\begin{figure}[H]
\centering
\includegraphics[width=0.75\textwidth]{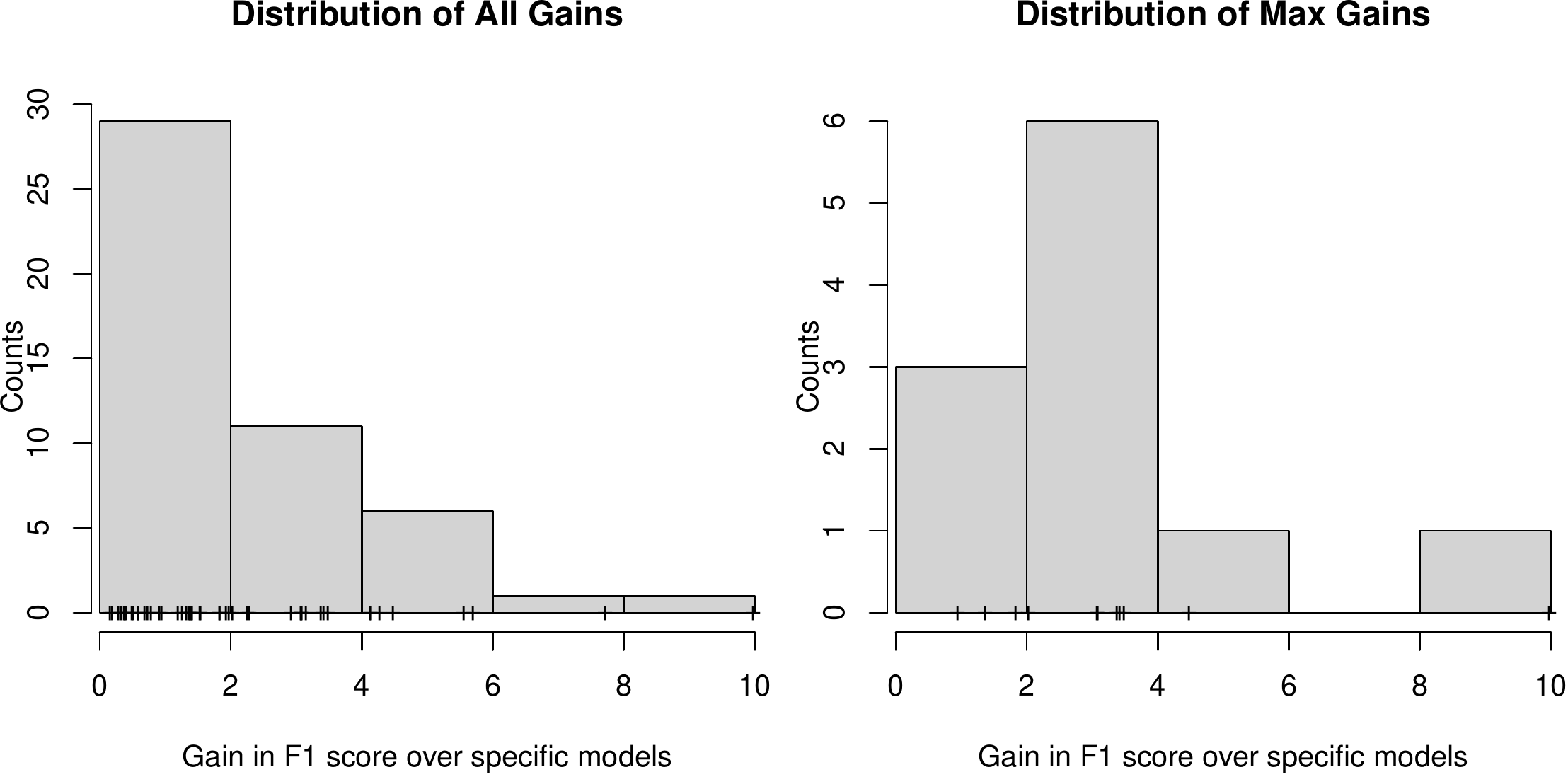}
\captionsetup{size=footnotesize}
\caption{Gains averaged over companies. Left: n=48, min=0.16, max=9.98, mean=2.22. Right: n=11, min=0.95, max=9.98, mean=3.37. \label{fig:gains}}
\end{figure}

The F1 scores averaged over all companies are shown in Table \ref{table:summ}.
Overall, the generic models bring improvement over the specific models for 73.3 \% of the domain-outcome combinations (11 out of 15).
As shown on the right of Fig. \ref{fig:gains}, maximum gains range from 0.95 (for electric T\&D-energy source) to 9.98 (for electric T\&D-injury type) with an average of 3.37.
Also, not only do the 11 best generic models outperform their specific counterparts with a comfortable margin, but they also generate finer-grained forecasts, which are much more useful in practice.

More specifically, generic models predict 2.26 additional categories on average, even up to 7 for construction-injury type (while still providing a gain of 3.48 in F1 score).
This is remarkable, considering that the more categories to be predicted, the more difficult the task (see \ref{sec:diff}). 

The construction and oil \& gas domains see gains for 3 outcomes out of 5, while on the electric T\&D domain, we observe improvement for every outcome.
Further, for the body part, injury type, and energy source outcomes, there is at least one generic model that outperforms its specific counterpart, on every domain, while the severity and accident type outcomes see improvements only on the electric T\&D domain.

However, it is important to note that even on those 4 domain-outcome combinations on which the generic models do not offer gains in predictive performance, it can still be desirable to use them in practice over the specific models, as they generate more informative forecasts, with 2 additional categories predicted, on average.

Overall, more than half of all F1 scores recorded for the generic models (79 out of 150, or 53\%) are greater or within two points of that of the specific models, while predicting 1.83 more categories on average.
And, as shown on the left of Fig. \ref{fig:gains}, the 48 generic models that outperform their specific counterparts bring on average an improvement of 2.22 in F1 score.

\subsection{Body part, injury type, and energy source}
Some of the greatest improvements are observed for injury type, where the best generic models provide large average gains of 3.48, 9.98, and 3.07, respectively on the construction, electric T\&D, and oil \& gas domains, while predicting on average 4.25 more categories than the company-specific models.
This large boost in performance is remarkable considering the significant increase in task difficulty.

Similarly, for energy source, the best generic models provide 4.48, 0.95, and 1.83 improvements in F1 score, while predicting 1.53 more categories on average; and for body part, the gains are 3.38, 3.42, and 1.37, with 0.17 more categories predicted.

\subsection{Severity and accident type}
For severity and accident type, the generic models outperform the company-specific ones on the electric T\&D domain, with 3.09 and 2.03 gains in F1 scores, while predicting 2 and 0.5 more categories on average.

On the construction and oil \& gas domains, the best generic models are between 2.4 and 6 points below the company-specific ones.
However, they still offer the benefit of predicting more categories (+1.6 on average).

\begin{table}[H]
\centering
\scalebox{0.85}{
\begin{tabular}{cccrcccccc}
\hline
& & & & \multicolumn{2}{c}{Construction} & \multicolumn{2}{c}{Electric T\&D} & \multicolumn{2}{c}{Oil \& Gas} \\
\hline
& & & & Full & Dom. & Full & Dom. & Full & Dom. \\
\hline
\hline
\multirow{9}{*}{\rotatebox[origin=c]{90}{{Severity}}} & \multirow{6}{*}{\rotatebox[origin=c]{90}{{F1}}} & SVM  & gen & 31.66 & 30.29 & 35.52 & 35.79 & 30.92 & 28.97 \\ 
\cline{3-10}
& & \multirow{2}{*}{RF} & gen & 27.26 & 31.04 & 33.37 & \textbf{41.28} & 23.48 & 26.08 \\
& & & ens & 30.33 & 31.82 & \textbf{41.17}$^{\dagger}$ & \textbf{43.88}$^{\dagger}$ & 28.98 & 30.76 \\  
\cline{3-10}
& & \multirow{2}{*}{XGB} & gen & 26.98 & 28.74 & 36.4 & 39.69$^{\star}$ & 24.25 & 24.19 \\  
& & & ens & 29.67 & 31.81 & 40.01$^{\star}$ & \textbf{40.95} & 29.14 & 31.44 \\ 
\cline{3-10}
& & & spec & \multicolumn{2}{c}{35.34$^{\dagger}$} & \multicolumn{2}{c}{40.79} & \multicolumn{2}{c}{33.86$^{\dagger}$} \\
 \cline{2-10}
& \multirow{3}{*}{\rotatebox[origin=c]{90}{{Count}}} & \multirow{2}{*}{\# categories} & spec & \multicolumn{2}{c}{3.5} & \multicolumn{2}{c}{3} & \multicolumn{2}{c}{3.5} \\ 
& &  & gen & 5 & 5 & 5 & 5 & 5 & 5 \\ 
\cline{3-10}
& &  \# datasets &  & 9 & 4 & 9 & 4 & 9 & 4 \\
\hline
\hline
\multirow{9}{*}{\rotatebox[origin=c]{90}{{Body Part}}} & \multirow{6}{*}{\rotatebox[origin=c]{90}{{F1}}} & SVM & gen & 25.51 & \textbf{31.32} & 26.68 & 25.24 & 23.65 & 24.66 \\ 
\cline{3-10}
& & \multirow{2}{*}{RF}  & gen & \textbf{34.41}$^{\dagger}$ & \textbf{31.81} & \textbf{34.94}$^{\dagger}$ & \textbf{33.05} & \textbf{30.22}$^{\dagger}$ & 28.85$^{\dagger}$ \\
& & & ens & 30.33$^{\star}$ & 30.39$^{\star}$ & \textbf{34.08} & 29.25 & 26.43 & 27.54$^{\star}$ \\  
\cline{3-10}
& & \multirow{2}{*}{XGB}  & gen & \textbf{32.86} & \textbf{32.23}$^{\dagger}$ & \textbf{33.76} & \textbf{35.21}$^{\dagger}$ & \textbf{29.22} & 28.59$^{\star}$ \\ 
& & & ens & 29.05$^{\star}$ &  29.2$^{\star}$ & \textbf{32.31} & \textbf{34.71} & 26.37 & 27.74$^{\star}$ \\   
\cline{3-10}
& & & spec & \multicolumn{2}{c}{31.03} & \multicolumn{2}{c}{31.79}  & \multicolumn{2}{c}{28.85} \\
\cline{2-10} 
& \multirow{3}{*}{\rotatebox[origin=c]{90}{{Count}}} & \multirow{2}{*}{\# categories} & spec & \multicolumn{2}{c}{6} & \multicolumn{2}{c}{5.5} & \multicolumn{2}{c}{6} \\   
& & & gen & 6 & 6 & 6 & 6 & 6 & 6 \\
\cline{3-10}
& &  \# datasets &  & 9 & 4 & 9 & 4 & 9 & 3 \\
\hline
\hline
\multirow{9}{*}{\rotatebox[origin=c]{90}{{Injury Type}}} & \multirow{6}{*}{\rotatebox[origin=c]{90}{{F1}}} & SVM & gen & 38.73 & 42.78$^{\star}$ & \textbf{53.7}$^{\dagger}$ & 42.91$^{\star}$ & 35.11$^{\star}$ & 33.89 \\ 
\cline{3-10}
& & \multirow{2}{*}{RF} & gen & 40.03 & 42.11$^{\star}$ & 42.44$^{\star}$ & 41.68$^{\star}$ & 32.15 & 32.82 \\ 
& & & ens & 41.75 & \textbf{45.46}$^{\dagger}$ & \textbf{49.42} & \textbf{45.10} & \textbf{38.65}$^{\dagger}$ & \textbf{38.97} \\ 
\cline{3-10}
& & \multirow{2}{*}{XGB} & gen & 36.76 & 42.55$^{\star}$ & 41.03 & 41.48 & 31.33 & 32.34 \\ 
& & & ens & \textbf{47.4}$^{\dagger}$ & \textbf{45.45} & \textbf{51.44} & \textbf{49.28}$^{\dagger}$ & \textbf{38.05} & \textbf{39.79}$^{\dagger}$ \\ 
\cline{3-10}
& & & spec & \multicolumn{2}{c}{43.92} & \multicolumn{2}{c}{43.72} & \multicolumn{2}{c}{36.72} \\ 
 \cline{2-10}
& \multirow{3}{*}{\rotatebox[origin=c]{90}{{Count}}} & \multirow{2}{*}{\# categories} & spec & \multicolumn{2}{c}{4} & \multicolumn{2}{c}{5.25} & \multicolumn{2}{c}{7} \\ 
& & & gen & 11 & 6 & 11 & 8 & 11 & 11 \\
\cline{3-10}
& &  \# datasets &  & 9 & 4 & 9 & 4 & 9 & 3 \\
\hline
\hline
\multirow{9}{*}{\rotatebox[origin=c]{90}{{Accident Type}}} & \multirow{6}{*}{\rotatebox[origin=c]{90}{{F1}}} & SVM & gen & 42.39 & 42.20 & 44.58 & 44.84 & 60.69 & 64.85 \\  
\cline{3-10}
& & \multirow{2}{*}{RF} & gen & 42.46 & 42.42 & \textbf{48.58}$^{\dagger}$ & \textbf{50.2}$^{\dagger}$ & 63.14 & 64.12 \\
& & & ens & 44.35 & 40.8 & 41.29 & 39.72 & 66.40 & 65.74 \\  
\cline{3-10}
& & \multirow{2}{*}{XGB} & gen & 48.27 & 49.21 & 47.80$^{\star}$ & \textbf{49.58} & 58.58 & 63.04\\  
& & & ens & 42.02 & 43.40 & 41.08 & 43.53 & 64.56 & 67.13 \\ 
\cline{3-10}
& & & spec & \multicolumn{2}{c}{54.98$^{\dagger}$} & \multicolumn{2}{c}{48.17} & \multicolumn{2}{c}{73.16$^{\dagger}$} \\
\cline{2-10}
& \multirow{3}{*}{\rotatebox[origin=c]{90}{{Count}}} & \multirow{2}{*}{\# categories} & spec & \multicolumn{2}{c}{3.5} & \multicolumn{2}{c}{4.5} & \multicolumn{2}{c}{2.67} \\ 
& & & gen & 5 & 5 & 5 & 5 & 5 & 4\\
\cline{3-10}
& &  \# datasets &  & 6 & 2 & 6 & 2 & 6 & 3 \\
\hline
\hline
\multirow{9}{*}{\rotatebox[origin=c]{90}{{Energy Source}}} & \multirow{6}{*}{\rotatebox[origin=c]{90}{{F1}}} & SVM & gen & \textbf{74.12}$^{\dagger}$ & \textbf{73.78} & 77.64$^{\star}$ & \textbf{78.87}$^{\dagger}$ & 69.54$^{\star}$ & 55.63 \\ 
\cline{3-10}
& & \multirow{2}{*}{RF} & gen & \textbf{72.71} & \textbf{73.91}$^{\dagger}$ & 77.12$^{\star}$ & \textbf{78.61} & \textbf{71.12} & \textbf{70.72}$^{\dagger}$ \\ 
& & & ens & 69.06$^{\star}$ & 69.64$^{\star}$ & 75.83 & 76.48$^{\star}$ & \textbf{70.58} & 69.12$^{\star}$ \\  
\cline{3-10}
& & \multirow{2}{*}{XGB} & gen & \textbf{73.77} & \textbf{71.04} & \textbf{78.83}$^{\dagger}$ & 75.61 & \textbf{72.22}$^{\dagger}$ & 70.35$^{\star}$ \\ 
& & & ens & 69.05$^{\star}$ & \textbf{70.23} & 76.47$^{\star}$ & 75.03 & \textbf{70.98} & 70.38$^{\star}$ \\  
\cline{3-10}
& & & spec & \multicolumn{2}{c}{69.64} & \multicolumn{2}{c}{77.92} & \multicolumn{2}{c}{70.39} \\
 \cline{2-10}
& \multirow{3}{*}{\rotatebox[origin=c]{90}{{Count}}} & \multirow{2}{*}{\# categories} & spec & \multicolumn{2}{c}{2.25} & \multicolumn{2}{c}{2.67} & \multicolumn{2}{c}{3} \\
& & & gen & 5 & 3 & 5 & 3 & 5 & 4 \\
\cline{3-10}
& & \# datasets &  & 8 & 4 & 8 & 3 & 8 & 2 \\
\hline
\end{tabular}
}
\captionsetup{size=scriptsize}
\caption{Results averaged over companies. $\dagger$: best of their sub-column.
\textbf{Bold}/$\star$: better than/within 2 pts of spec.
Full: full generic model (one per outcome, same across domains).
Dom.: per-domain generic model (one per outcome per domain).
Gen/spec: generic/specific.
Ens: ensemble thereof.
\# datasets: number of company datasets forming the generic dataset.
Note: for the same outcome, \# categories and \# datasets are the same for Full across domains, we repeat them only to ease comparison.}\label{table:summ}
\end{table}

\subsection{Full vs. per-domain}
In what follows, we refer to the full and per-domain models and their ensemble versions.
When considering full generic models, the average improvement in F1-score over the specific models is 2.85 and there are 2.44 additional categories predicted (min=0, max=7), while when considering per-domain generic models, the average improvement is 2.57 and 1.38 additional categories are predicted (min=0, max=4).
The per-domain models reach a higher max score than the full models on 9 combinations out of 15 (60\%), and in 5 out of 11 (45\%) when the specific models are outperformed.
The full and per-domain models outperform the specific models on the same 11 domain-outcome combinations.

So, in terms of performance, there is no clear winner.
However, since the full generic models predict more categories, and are also simpler conceptually (just one model per outcome), full models seem like the way to go.
This conclusion however will need to be validated when more datasets are available for each domain.
One thing to note, however, is that specific models may still be desirable in the context of model stacking, as covered next.

\subsection{Generic vs. ensemble (generic + specific)}
The transfer learning-like stacking approach, i.e., combining the predictions of the generic and specific models, boosts performance over the generic models (both full and per-domain) on all domains for the severity and injury type outcomes, in some cases for accident type, and nowhere for body part and energy source.

For severity, the average gains are of 3.93, and range from 0.78 to an impressive 7.8 (for electric T\&D-full-RF).
Results are even more impressive for injury type.
Gains range from 1.72 to 10.64 (for construction-full-XGB), with a high average of 6.17.

It is interesting to note that for severity and injury type, very few of the generic models outperform the specific models in the first place, and it is only by combining their predictions with that of the specific models that absolute best performance can be reached, on the electrical domain for severity, and on all domains for injury type.

We also observe that conversely, for body part and energy source, where model stacking does not bring additional skill, the generic models are stronger than the specific models in the first place.

All in all, these results may suggest that ensembling only works when the generic models are not already better than the specific models.
However, this rule does not hold everywhere (e.g., construction-accident type-XGB), so additional data, experiments and results will be necessary to draw any general conclusion here.

\subsection{Quantity vs. relevance}
\begin{figure}[H]
\centering

\begin{minipage}{0.48\linewidth}\centering
\includegraphics[width=0.9\textwidth]{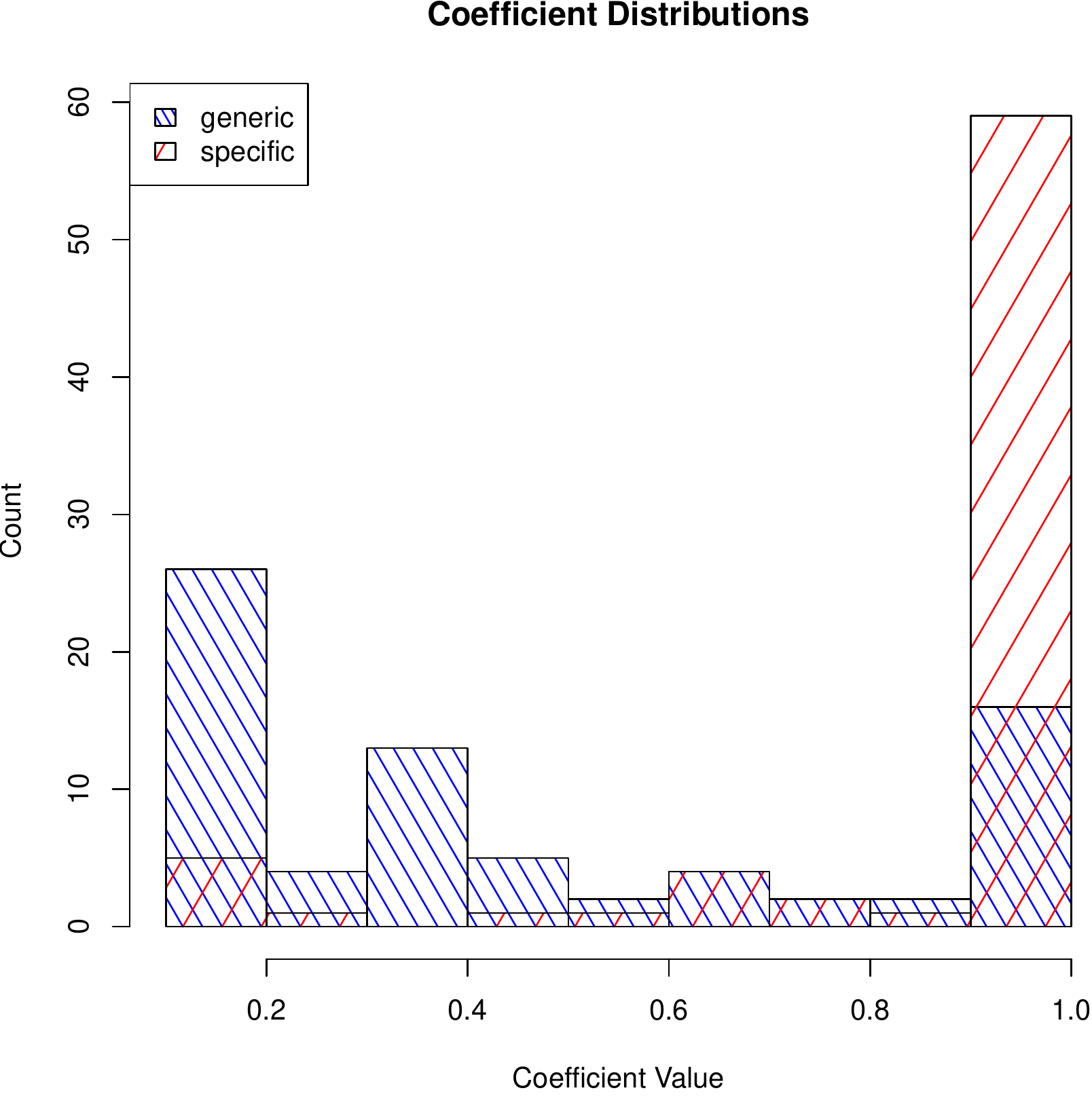}
\end{minipage}
\begin{minipage}{0.48\linewidth}\centering
\includegraphics[width=0.9\textwidth]{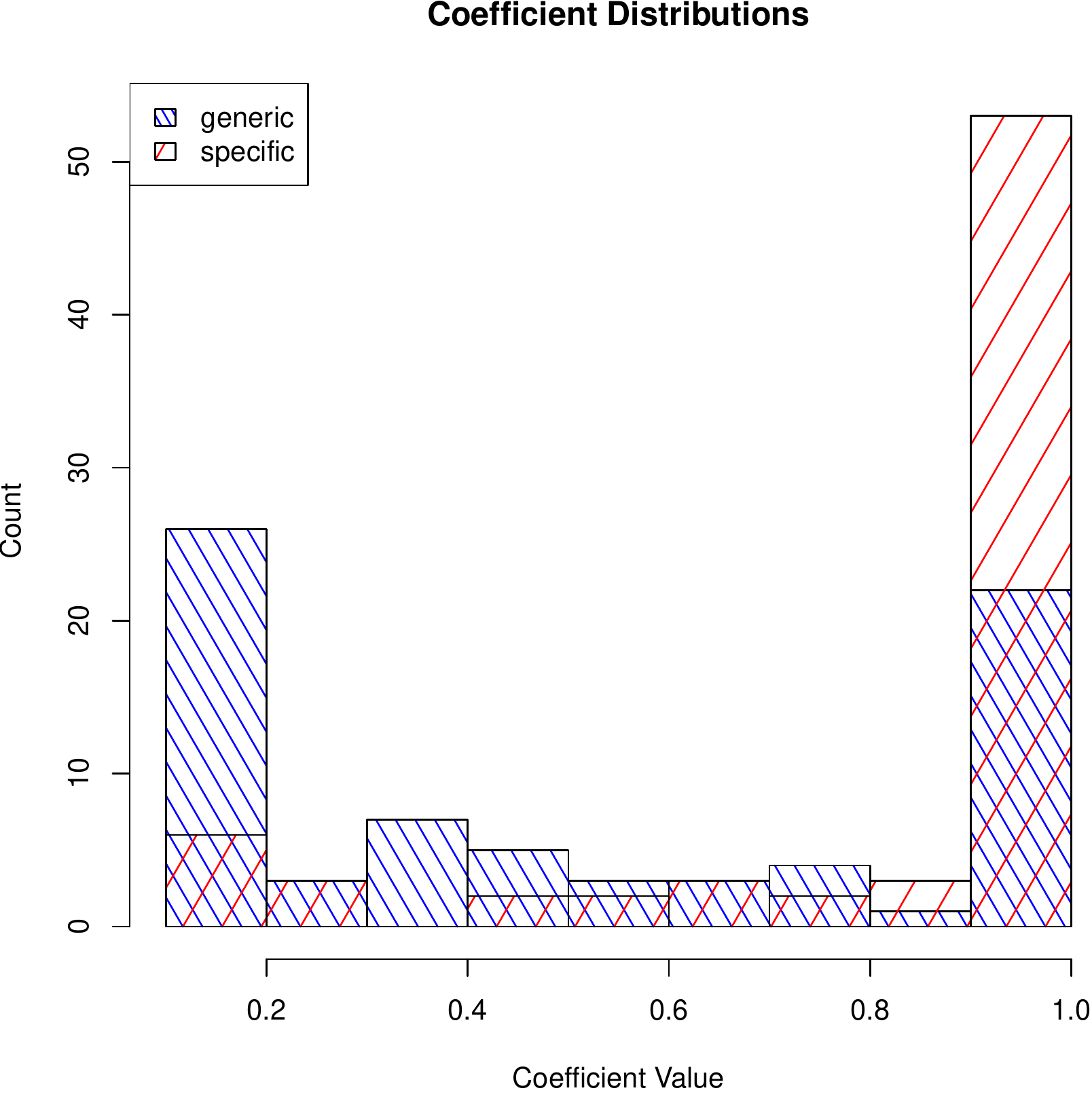}
\end{minipage}

\captionsetup{size=footnotesize}
\caption{Distributions of the best coefficient values $a$ (generic) and $b$ (specific).
Left: full. Right: per-domain. \label{fig:coeff_distr}}
\end{figure}

As far as whether more data or more \textit{relevant} data is best, Fig. \ref{fig:coeff_distr} shows the distributions of the best $a$ and $b$ coefficients as determined on the validation sets.
It tends to indicate that, on average, the best tradeoff involves anywhere from a little bit to a lot of generic model (anywhere in the [0.1,1] range, with peaks towards [0.1,0.2] and [0.9,1]), but almost always a lot of specific model (between 0.9 and 1).
In other words, data relevance always seems important, while the contribution of data quantity fluctuates.
However, this is only a general trend.
As can be seen in the detailed results per company (\ref{sec:full_details} and \ref{sec:generic_details}), in some cases, the contribution of the generic model is more important than that of the specific model, e.g., (1,0.6) for Company6-XGB in the first table of \ref{sec:generic_details}.

\subsection{Best model type}
For the full generic models, the best algorithm is RF (6 domain-outcome combinations over 15), followed by SVM (5/15) and XGB (4/15).
When stacked with the specific model, RF reaches best performance in 10 out of 15 combinations.

When considering the per-domain generic models, SVM obtains the best score 7 times out of 15, followed by RF (5/15) and XGB (3/5).
However, when used in the ensemble, XGB is the best (10/15).

RF and XGB are better choices than SVM as they consistently top the scores and can be used in ensembles.
In terms of performance though, there is no clear winner between the two.
One or the other could be used interchangeably.
However, XGBoost is superior in practice as far as deployment is concerned, as the Random Forest models take a lot of disk space, even after applying some compression tricks.

\section{Conclusion}
We showed that generic models provide consistent and large improvements over company-specific models.
Moreover, generic models issue finer-grained forecasts that are more useful in practice, as they predict more categories of each safety outcome.

Generic models remove the needs for training company-specific models, saving a lot of time and resources, and give small companies, whose accident datasets are too limited to train their own models, access to safety outcome predictions.

Per-domain generic models (trained on data from a specific industry sector) are not always better than full generic models (trained on all data).
Ensembling generic and specific models is often very beneficial.
Therefore, it might still be worth training specific models to combine their predictions with that of the generic models.
If specific models are already in use, combining them with the generic models may provide a boost in performance.

The forecasts are in essence clear and direct information that can be accessed via a user interface (as a desktop or mobile webpage or application), or via an API for integration into any existing ecosystem.
In each case, the only input required is a set of attributes, and the output are probabilities for each category of each outcome.

By learning lessons from a pool of datasets whose accumulated experience far exceeds that of any single company, and making these lessons easily accessible, generic models tackle the holy grail of safety cross-organizational learning and dissemination in the construction industry.

\section{Acknowledgements} 
We thank the Nvidia corporation for donating the Titan V GPU that was used in this research, as part of their GPU grant program.

\section{References}

\footnotesize

\bibliographystyle{elsarticle-num} 
\bibliography{bib}

\begin{thebibliography}{10}
\expandafter\ifx\csname url\endcsname\relax
  \def\url#1{\texttt{#1}}\fi
\expandafter\ifx\csname urlprefix\endcsname\relax\def\urlprefix{URL }\fi
\expandafter\ifx\csname href\endcsname\relax
  \def\href#1#2{#2} \def\path#1{#1}\fi

\bibitem{desvignes2014requisite}
M.~Desvignes, \href{https://scholar.colorado.edu/downloads/0r967398g}{Requisite
  empirical risk data for integration of safety with advanced technologies and
  intelligent systems}, Ph.D. thesis, University of Colorado at Boulder (2014).
\newline\urlprefix\url{https://scholar.colorado.edu/downloads/0r967398g}

\bibitem{villanova2014attribute}
M.~P. Villanova,
  \href{https://scholar.colorado.edu/downloads/jd472w76t}{Attribute-based risk
  model for assessing risk to industrial construction tasks}, Ph.D. thesis,
  University of Colorado at Boulder (2014).
\newline\urlprefix\url{https://scholar.colorado.edu/downloads/jd472w76t}

\bibitem{tixier2016automated}
A.~J.-P. Tixier, M.~R. Hallowell, B.~Rajagopalan, D.~Bowman, Automated content
  analysis for construction safety: a natural language processing system to
  extract precursors and outcomes from unstructured injury reports, Automation
  in Construction 62 (2016) 45--56.
\newblock \href {http://dx.doi.org/10.1016/j.autcon.2015.11.001}
  {\path{doi:10.1016/j.autcon.2015.11.001}}.

\bibitem{tixier2016application}
A.~J.-P. Tixier, M.~R. Hallowell, B.~Rajagopalan, D.~Bowman, Application of
  machine learning to construction injury prediction, Automation in
  Construction 69 (2016) 102--114.
\newblock \href {http://dx.doi.org/10.1016/j.autcon.2016.05.016}
  {\path{doi:10.1016/j.autcon.2016.05.016}}.

\bibitem{baker2020ai}
H.~Baker, M.~R. Hallowell, A.~J.-P. Tixier, Ai-based prediction of independent
  construction safety outcomes from universal attributes, Automation in
  Construction 118 (2020) 103146.

\bibitem{tixier2017clash}
A.~J.-P. Tixier, M.~R. Hallowell, B.~Rajagopalan, D.~Bowman, Construction
  safety clash detection: identifying safety incompatibilities among
  fundamental attributes using data mining, Automation in Construction 74
  (2017) 39--54.

\bibitem{tixier2017construction}
A.~J.-P. Tixier, M.~R. Hallowell, B.~Rajagopalan, Construction safety risk
  modeling and simulation, Risk Analysis 37~(10) (2017) 1917--1935.
\newblock \href {http://dx.doi.org/10.1111/risa.12772}
  {\path{doi:10.1111/risa.12772}}.

\bibitem{tanguy2016natural}
L.~Tanguy, N.~Tulechki, A.~Urieli, E.~Hermann, C.~Raynal, Natural language
  processing for aviation safety reports: From classification to interactive
  analysis, Computers in Industry 78 (2016) 80--95.

\bibitem{sepeda2006lessons}
A.~L. Sepeda, Lessons learned from process incident databases and the process
  safety incident database (psid) approach sponsored by the center for chemical
  process safety, Journal of hazardous materials 130~(1-2) (2006) 9--14.

\bibitem{le2014social}
Q.~T. Le, D.~Y. Lee, C.~S. Park, A social network system for sharing
  construction safety and health knowledge, Automation in Construction 46
  (2014) 30--37.

\bibitem{pedro2022data}
A.~Pedro, A.-T. Pham-Hang, P.~T. Nguyen, H.~C. Pham, Data-driven construction
  safety information sharing system based on linked data, ontologies, and
  knowledge graph technologies, International journal of environmental research
  and public health 19~(2) (2022) 794.

\bibitem{edwin2022sharing}
K.~W. Edwin, Sharing incident experiences: a roadmap towards collective safety
  information in the norwegian construction industry, International Journal of
  Occupational Safety and Ergonomics (2022) 1--11.

\bibitem{wasilkiewicz2018information}
K.~Wasilkiewicz, Information flow and knowledge transfer of accident
  investigation results in the norwegian construction industry, in: Safety and
  Reliability--Safe Societies in a Changing World, CRC Press, 2018, pp.
  2855--2862.

\bibitem{breiman2001random}
L.~Breiman, Random forests, Machine Learning 45~(1) (2001) 5--32.

\bibitem{chen2016xgboost}
T.~Chen, C.~Guestrin, Xgboost: A scalable tree boosting system, in: Proceedings
  of the 22nd ACM SIGKDD International Conference on Knowledge Discovery and
  Data Mining, ACM, 2016, pp. 785--794.

\bibitem{boser1992training}
B.~E. Boser, I.~M. Guyon, V.~N. Vapnik, A training algorithm for optimal margin
  classifiers, in: ACM Proceedings of the Fifth Annual Workshop on
  Computational learning theory, 1992, pp. 144--152.
\newblock \href {http://dx.doi.org/10.1145/130385.130401}
  {\path{doi:10.1145/130385.130401}}.

\bibitem{krizhevsky2012imagenet}
A.~Krizhevsky, I.~Sutskever, G.~E. Hinton, Imagenet classification with deep
  convolutional neural networks, in: Advances in neural information processing
  systems, 2012, pp. 1097--1105.

\bibitem{radford2018improving}
A.~Radford, K.~Narasimhan, T.~Salimans, I.~Sutskever, Improving language
  understanding by generative pre-training, URL https://s3-us-west-2.
  amazonaws. com/openai-assets/researchcovers/languageunsupervised/language
  understanding paper. pdf.

\bibitem{devlin2018bert}
J.~Devlin, M.-W. Chang, K.~Lee, K.~Toutanova, Bert: Pre-training of deep
  bidirectional transformers for language understanding, arXiv preprint
  arXiv:1810.04805.

\bibitem{lewis2019bart}
M.~Lewis, Y.~Liu, N.~Goyal, M.~Ghazvininejad, A.~Mohamed, O.~Levy, V.~Stoyanov,
  L.~Zettlemoyer, Bart: Denoising sequence-to-sequence pre-training for natural
  language generation, translation, and comprehension, arXiv preprint
  arXiv:1910.13461.

\bibitem{eddine2020barthez}
M.~K. Eddine, A.~J.-P. Tixier, M.~Vazirgiannis, Barthez: a skilled pretrained
  french sequence-to-sequence model, arXiv preprint arXiv:2010.12321.

\bibitem{r_language}
{R Core Team}, \href{https://www.R-project.org/}{R: A Language and Environment
  for Statistical Computing}, R Foundation for Statistical Computing, Vienna,
  Austria (2022).
\newline\urlprefix\url{https://www.R-project.org/}

\bibitem{scikit-learn}
F.~Pedregosa, G.~Varoquaux, A.~Gramfort, V.~Michel, B.~Thirion, O.~Grisel,
  M.~Blondel, P.~Prettenhofer, R.~Weiss, V.~Dubourg, J.~Vanderplas, A.~Passos,
  D.~Cournapeau, M.~Brucher, M.~Perrot, E.~Duchesnay, Scikit-learn: Machine
  learning in {P}ython, Journal of Machine Learning Research 12 (2011)
  2825--2830.

\end{thebibliography}

\normalsize

\vspace{1.5cm}

{\LARGE\textbf{Appendices}}

\appendix

\section{Attribute List}\label{sec:atts}

\begin{table}[H]
\small
\centering
\scalebox{0.8}{
\begin{tabular}{llll}
  \hline
adverse low temps & fuses$^\star$ & machinery & spark \\ 
  bolt & grinding & manlift & splinter/sliver \\ 
  breaker$^\star$ & grout & mud & spool \\ 
  cable & guardrail/handrail & nail & stairs \\ 
  cable tray & hammer & no/improper PPE & steel/steel sections \\ 
  chipping & hand size pieces & object at height & stripping \\ 
  cleaning & hazardous substance & object on the floor & stud \\ 
  clearance$^\star$ & heat source/high temps & piping & switch/switching$^\star$ \\ 
  concrete & heater$^\star$ & pole$^\star$ & tank \\ 
  concrete liquid & heavy material/tool & pontoon & transformer$^\star$ \\ 
  conduit & heavy vehicle & poor housekeeping & uneven surface \\ 
  confined work space & hose & poor visibility & unpowered tool \\ 
  congested work space & improper body position & powered tool & unpowered transporter \\ 
  crane & improper procedure/inattention & rebar & unstable support/surface \\ 
  door & improper security of materials & relay$^\star$ & valve \\ 
  drill & improper security of tools & repetitive motion & vault$^\star$ \\ 
  dunnage & insect/animal & scaffold & welding \\ 
  electricity & job trailer & screw & wind \\ 
  exiting & ladder & sharp edge & wire \\ 
  fan$^\star$ & lifting/pulling/manipulating & slag & working at height \\ 
  fatigued dizzy & light vehicle & slippery surface & working below elev wksp/mat \\ 
  forklift & LOTO/labeling$^\star$ & small particle & working overhead \\ 
  formwork & lumber & soffit & wrench \\ 
   \hline
\end{tabular}
}
\caption{92 attributes used in this study. LOTO: lockout-tagout. PPE: personal protective equipment. $\star$: eleven new attributes added since \cite{tixier2016application,baker2020ai}.}
\label{table:atts}
\end{table}

\section{Detailed split counts}

\begin{table}[H]
\centering
\begin{minipage}{0.48\linewidth}\centering
\scalebox{0.795}{
\begin{tabular}{crrrrr}
\hline
\multicolumn{6}{c}{Severity} \\
  \hline
& & Train & $w$ & Val & Test  \\ 
  \hline
\multirow{6}{*}{\rotatebox[origin=c]{90}{{Construction}}}
&  report-only & 917 & 8.2 & 226 & 283 \\ 
& 1st aid & 7486 & 1.0 & 1876 & 2369 \\ 
&  medical & 470 & 15.9 & 114 & 140 \\ 
&  recordable & 147 & 50.9 &  28 &  42 \\ 
&  lost time & 960 & 7.8 & 250 & 285 \\ 
\cline{2-6}
&  total & \textbf{9980} &  & \textbf{2494} & \textbf{3119} \\ 
   \hline
\multirow{6}{*}{\rotatebox[origin=c]{90}{{Electric T\&D}}}
&  report-only & 2392 & 1.2 & 576 & 712 \\ 
&   1st aid & 2809 & 1.0 & 736 & 905 \\ 
&  medical & 554 & 5.1 & 140 & 162 \\ 
&  recordable & 310 & 9.1 &  74 & 101 \\ 
&  lost time & 607 & 4.6 & 143 & 205 \\ 
\cline{2-6}
&  total & \textbf{6672} &  & \textbf{1669} & \textbf{2085} \\
    \hline
\multirow{6}{*}{\rotatebox[origin=c]{90}{{Oil \& Gas}}}
&  report-only & 929 & 14.8 & 244 & 279 \\ 
& 1st aid & 13766 & 1.0 & 3405 & 4279 \\ 
&  medical & 1919 & 7.2 & 489 & 618 \\
&  recordable & 152 & 90.6 &  42 &  52 \\ 
&  lost time & 1615 & 8.5 & 415 & 516 \\ 
\cline{2-6}
&  total & \textbf{18381} &  & \textbf{4595} & \textbf{5744} \\
    \hline
\multirow{3}{*}{\rotatebox[origin=c]{90}{{\scriptsize Corporate}}}
 & report-only &  97 & 3.3 &  31 &  22 \\ 
& 1st aid & 321 & 1.0 &  74 & 109 \\ 
\cline{2-6}
 & total & \textbf{418} &  & \textbf{105} & \textbf{131} \\
    \hline
\multirow{6}{*}{\rotatebox[origin=c]{90}{{Full}}}
&  report-only & 4335 & 5.6 & 1077 & 1296 \\
& 1st aid & 24382 & 1.0 & 6091 & 7662 \\
&  medical & 2943 & 8.3 & 743 & 920 \\ 
&  recordable & 609 & 40.0 & 144 & 195 \\ 
&  lost time & 3182 & 7.7 & 808 & 1006 \\ 
\cline{2-6}
&  total & \textbf{35451} &  & \textbf{8863} & \textbf{11079} \\ 
   \hline
\end{tabular}
}
\end{minipage}
\begin{minipage}{0.48\linewidth}\centering

\scalebox{0.785}{

\begin{tabular}{crrrrr}
\hline
\multicolumn{6}{c}{Body Part} \\
  \hline
& & Train & $w$ & Val & Test \\ 
  \hline
\multirow{7}{*}{\rotatebox[origin=c]{90}{{Construction}}}
& arm & 1059 & 2.6 & 285 & 338 \\ 
&  foot & 694 & 3.9 & 167 & 232 \\ 
&  hand & 2732 & 1.0 & 701 & 864 \\ 
&  head & 1682 & 1.6 & 394 & 494 \\ 
&  leg & 958 & 2.9 & 262 & 307 \\ 
&  trunk & 1084 & 2.5 & 243 & 330 \\ 
\cline{2-6}
&  total & \textbf{8209} &  & \textbf{2052} & \textbf{2565} \\ 
   \hline
   \multirow{7}{*}{\rotatebox[origin=c]{90}{{Electric T\&D}}}
& arm & 1061 & 1.4 & 274 & 319 \\ 
&  foot & 372 & 4.0 &  89 & 135 \\ 
&  hand & 1473 & 1.0 & 368 & 452 \\ 
&  head & 1246 & 1.2 & 318 & 403 \\ 
&  leg & 1084 & 1.4 & 251 & 307 \\ 
&  trunk & 800 & 1.8 & 208 & 269 \\ 
\cline{2-6}
&  total & \textbf{6036} &  & \textbf{1508} & \textbf{1885} \\ 
   \hline
   \multirow{7}{*}{\rotatebox[origin=c]{90}{{Oil \& Gas}}}
& arm & 1445 & 3.9 & 386 & 477 \\ 
&  foot & 1741 & 3.2 & 421 & 568 \\ 
&  hand & 5586 & 1.0 & 1385 & 1740 \\ 
&  head & 3514 & 1.6 & 887 & 1088 \\ 
&  leg & 2053 & 2.7 & 498 & 596 \\ 
&  trunk & 1449 & 3.9 & 370 & 464 \\ 
\cline{2-6}
&  total & \textbf{15788} &  & \textbf{3947} & \textbf{4933} \\ 
   \hline
   \multirow{7}{*}{\rotatebox[origin=c]{90}{{Full}}}
& arm & 3565 & 2.7 & 945 & 1134 \\ 
&  foot & 2807 & 3.5 & 677 & 935 \\ 
&  hand & 9791 & 1.0 & 2454 & 3056 \\ 
&  head & 6442 & 1.5 & 1599 & 1985 \\ 
&  leg & 4095 & 2.4 & 1011 & 1210 \\ 
&  trunk & 3333 & 2.9 & 821 & 1063 \\ 
   \cline{2-6}
&  total & \textbf{30033} &  & \textbf{7507} & \textbf{9383} \\ 
   \hline
\end{tabular}
}
\end{minipage}

\vspace{1cm}

\begin{minipage}{0.477\linewidth}\centering

\scalebox{0.785}{
\begin{tabular}{crrrrr}
\hline
\multicolumn{6}{c}{Accident Type} \\
  \hline
& & Train & $w$ & Val & Test \\ 
  \hline
  \multirow{6}{*}{\rotatebox[origin=c]{90}{{Construction}}}
& caught & 396 & 2.3 & 105 & 137 \\ 
&  exposure & 119 & 7.8 &  38 &  40 \\ 
&  fall & 803 & 1.2 & 200 & 243 \\ 
&  overexertion & 492 & 1.9 & 128 & 160 \\ 
&  struck & 930 & 1.0 & 214 & 276 \\ 
   \cline{2-6}
&  total & \textbf{2740} &  & \textbf{685} & \textbf{856} \\ 
   \hline
     \multirow{6}{*}{\rotatebox[origin=c]{90}{{Electric T\&D}}}
&   caught & 207 & 2.2 &  55 &  62 \\ 
&  exposure & 454 & 1.0 & 123 & 142 \\ 
&  fall & 403 & 1.1 & 102 & 143 \\ 
&  overexertion & 288 & 1.6 &  51 &  65 \\ 
&  struck & 248 & 1.8 &  69 &  88 \\ 
   \cline{2-6}
&  total & \textbf{1600} &  & \textbf{400} & \textbf{500} \\ 
   \hline
    \multirow{5}{*}{\rotatebox[origin=c]{90}{{Oil \& Gas}}}
&   caught & 198 & 7.7 &  43 &  53 \\ 
&  exposure & 526 & 2.9 & 127 & 184 \\ 
&  fall & 1527 & 1.0 & 393 & 463 \\ 
&  struck & 659 & 2.3 & 165 & 210 \\ 
   \cline{2-6}
&  total & \textbf{2910} &  & \textbf{728} & \textbf{910} \\ 
   \hline
   \multirow{6}{*}{\rotatebox[origin=c]{90}{{Full}}}
&   caught & 801 & 3.4 & 203 & 252 \\ 
&  exposure & 1099 & 2.5 & 288 & 366 \\ 
&  fall & 2733 & 1.0 & 695 & 849 \\ 
&  overexertion & 780 & 3.5 & 179 & 225 \\ 
&  struck & 1837 & 1.5 & 448 & 574 \\ 
   \cline{2-6}
&  total & \textbf{7250} &  & \textbf{1813} & \textbf{2266} \\ 
   \hline
\end{tabular}
}
\end{minipage}
\begin{minipage}{0.515\linewidth}\centering

\scalebox{0.825}{
\begin{tabular}{crrrrr}
\hline
\multicolumn{6}{c}{Energy Source} \\
  \hline
& & Train & $w$ & Val & Test \\ 
  \hline
  \multirow{4}{*}{\rotatebox[origin=c]{90}{{\scriptsize Construction}}}
& chemical &  76 & 42.7 &  21 &  14 \\ 
& gravity & 1551 & 2.1 & 405 & 479 \\ 
&  motion & 3248 & 1.0 & 792 & 1031 \\ 
   \cline{2-6}
&  total & \textbf{4875} &  & \textbf{1218} & \textbf{1524} \\ 
   \hline
 \multirow{4}{*}{\rotatebox[origin=c]{90}{{Electric}}}
&  biological & 221 & 7.6 &  52 &  88 \\ 
&  gravity & 733 & 2.3 & 179 & 230 \\ 
&  motion & 1683 & 1.0 & 429 & 507 \\ 
   \cline{2-6}
&  total & \textbf{2637} &  & \textbf{660} & \textbf{825} \\ 
   \hline
 \multirow{5}{*}{\rotatebox[origin=c]{90}{{Oil \& Gas}}}
&  chemical &  70 & 21.2 &  13 &  21 \\ 
&  gravity & 1485 & 1.0 & 361 & 448 \\ 
&  motion & 914 & 1.6 & 246 & 300 \\ 
&  thermal & 131 & 11.3 &  30 &  44 \\ 
   \cline{2-6}
&  total & \textbf{2600} &  & \textbf{650} & \textbf{813} \\ 
   \hline
\multirow{6}{*}{\rotatebox[origin=c]{90}{{Full}}}\textbf{}
&   biological & 221 & 26.4 &  52 &  88 \\ 
&  chemical & 146 & 40.0 &  34 &  35 \\ 
&  gravity & 3769 & 1.6 & 945 & 1157 \\ 
&  motion & 5845 & 1.0 & 1467 & 1838 \\ 
&  thermal & 131 & 44.6 &  30 &  44 \\
   \cline{2-6}
&  total & \textbf{10112} &  & \textbf{2528} & \textbf{3162} \\ 
   \hline
\end{tabular}
}

\end{minipage}

\caption{Split counts (1/2). $w$: training weights.}\label{table:splits}
\end{table}

\begin{table}[ht]
\centering
\scalebox{0.7175}{
\begin{tabular}{crrrrr}
\hline
\multicolumn{6}{c}{Injury Type} \\
  \hline
& & Train & $w$ & Val & Test \\
  \hline
\multirow{7}{*}{\rotatebox[origin=c]{90}{{Construction}}}
& contusion & 728 & 3.6 & 185 & 229 \\ 
&  cut & 2644 & 1.0 & 682 & 795 \\ 
&  fob & 399 & 6.6 &  84 & 118 \\ 
&  fracture & 100 & 26.4 &  24 &  39 \\ 
&  pinch & 267 & 9.9 &  90 &  97 \\ 
&  strain & 2129 & 1.2 & 501 & 680 \\ 
\cline{2-6}
&  total & \textbf{6267} &  & \textbf{1566} & \textbf{1958} \\ 
   \hline
\multirow{9}{*}{\rotatebox[origin=c]{90}{{Electric T\&D}}}
&   bite & 129 & 12.3 &  35 &  42 \\ 
&  burn &  75 & 21.2 &  14 &  21 \\ 
&  contusion & 861 & 1.8 & 216 & 277 \\ 
&  cut & 1305 & 1.2 & 330 & 400 \\ 
&  fob & 209 & 7.6 &  46 &  69 \\ 
&  fracture & 176 & 9.0 &  39 &  53 \\ 
&  irritation & 420 & 3.8 & 101 & 141 \\ 
&  strain & 1589 & 1.0 & 410 & 486 \\ 
\cline{2-6}
&  total & \textbf{4764} &  & \textbf{1191} & \textbf{1489} \\ 
   \hline
   \multirow{12}{*}{\rotatebox[origin=c]{90}{{Oil \& Gas}}}
&   bite & 168 & 27.6 &  39 &  52 \\ 
&  burn & 572 & 8.1 & 150 & 179 \\ 
&  contusion & 3587 & 1.30 & 848 & 1091 \\ 
&  cut & 4638 & 1.0 & 1160 & 1509 \\ 
&  exhaustion &  75 & 61.8 &  24 &  25 \\ 
&  fob & 1440 & 3.2 & 381 & 455 \\ 
&  fracture & 622 & 7.5 & 160 & 199 \\ 
&  irritation & 127 & 36.5 &  37 &  42 \\ 
&  pain & 704 & 6.6 & 176 & 215 \\ 
&  pinch & 720 & 6.4 & 181 & 231 \\ 
&  strain & 2307 & 2.0 & 584 & 677 \\ 
\cline{2-6}
&  total & \textbf{14960} &  & \textbf{3740} & \textbf{4675} \\ 
   \hline
   \multirow{12}{*}{\rotatebox[origin=c]{90}{{Full}}}
&   bite & 297 & 28.9 &  74 &  94 \\ 
&  burn & 647 & 13.3 & 164 & 200 \\ 
&  contusion & 5176 & 1.7 & 1249 & 1597 \\ 
&  cut & 8587 & 1.0 & 2172 & 2704 \\ 
&  exhaustion &  75 & 114.5 &  24 &  25 \\ 
&  fob & 2048 & 4.2 & 511 & 642 \\ 
&  fracture & 898 & 9.6 & 223 & 291 \\ 
&  irritation & 547 & 15.7 & 138 & 183 \\ 
&  pain & 704 & 12.2 & 176 & 215 \\ 
&  pinch & 987 & 8.7 & 271 & 328 \\ 
&  strain & 6025 & 1.4 & 1495 & 1843 \\ 
\cline{2-6}
&  total & \textbf{25991} &  & \textbf{6497} & \textbf{8122} \\ 
   \hline
\end{tabular}
}
\caption{Split counts (2/2). $w$: training weights.}\label{table:splits_1}
\end{table}

\section{Hyperparameter Optimization Details}\label{sec:hyper}

For Random Forest\footnote{\href{https://scikit-learn.org/stable/modules/generated/sklearn.ensemble.RandomForestClassifier.html}{https://scikit-learn.org/stable/modules/generated/sklearn.ensemble.RandomForestClassifier.html}}, we searched the number of trees (\texttt{ntree} parameter, from 100 to 1600 with steps of 100), the number of variables to try when making each split (\texttt{mtry}, from 5 to 45 with steps of 5), and the leaf size (\texttt{nodesize}, 1, 2, 5, 10, 25, and 50).

For XGBoost\footnote{ \href{https://xgboost.readthedocs.io/en/latest/parameter.html}{https://xgboost.readthedocs.io/en/latest/parameter.html}}, we searched the maximum depth of a tree in the sequence (\texttt{max\_depth}, from 3 to 6 with steps of 1), the learning rate (\texttt{learning\_rate}, 0.01, 0.05, and 0.1), the minimum leaf size (\texttt{min\_child\_weight}, 1, 3, 5, and 10), the percentage of training instances to be used in building each tree (\texttt{subsample}, 0.3, 0.5, 0.7, and 1) , and the percentage of predictors to be considered in making each split of a given tree (\texttt{colsample\_bylevel}, 0.3, 0.5, 0.7, and 1).
The number of trees in the sequence (\texttt{ntrees}) was set to 2000.
The loss was the multinomial one.
Finally, for the SVM model, we optimized the C parameter (\texttt{C}, $10^x$ with $x$ taking 3000 evenly spaced values in $[-9,9]$).

\section{Illustration of Task Difficulty vs. Number of Categories}\label{sec:diff}

\begin{figure}[h]
\centering
\includegraphics[width=0.46\textwidth]{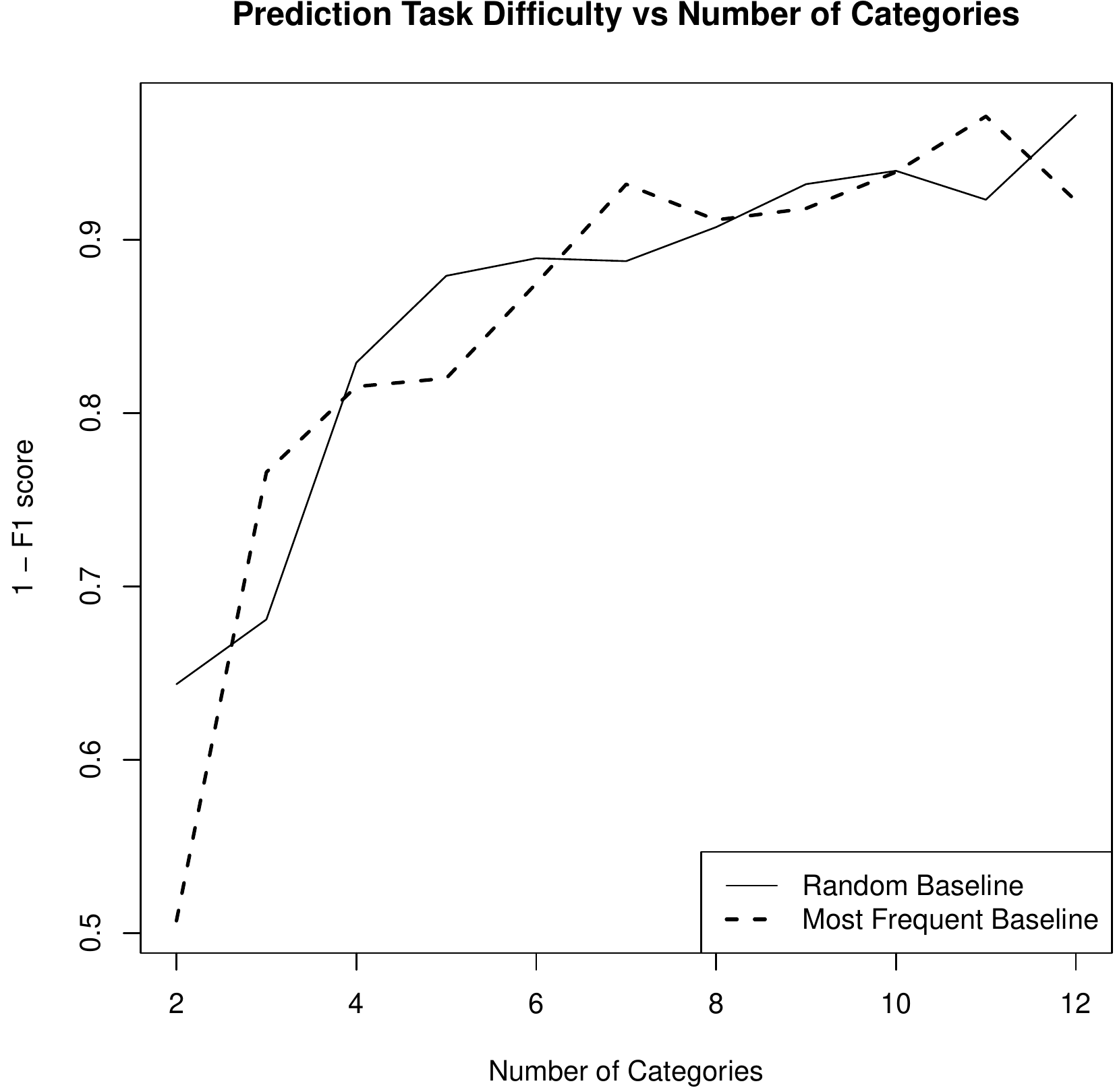}
\captionsetup{size=footnotesize}
\caption{\label{fig:diff}}
\end{figure}

To illustrate how the prediction task gets more and more difficult as the number of categories increases, we designed a synthetic example in which $10^5$ observations were drawn from an increasing number of categories (2 to 12).
Class imbalance was simulated by drawing from the categories with probabilities following the lognormal distribution (mean=0, sd=2).
We considered two baselines: a random baseline, that predicts categories uniformly at random, and a most frequent baseline, which always returns the most frequent category.
Our proxy for difficulty was one minus the F1 score of the baselines.
In other words, the less well the baselines are doing, the more difficult the task.
We can see on Fig. \ref{fig:diff} that the task difficulty rapidly increases with the number of categories, and that going from 2 to 6 categories almost makes the task twice as hard.

\section{Per-Company Results for the Full Generic Models}\label{sec:full_details}

Note: the ensemble (``ens'') rows are left blank whenever the specific model is a SVM, as we could not use ensembling in this case (the forecast of the SVM is not probabilistic).

\subsection{Severity}

\begin{table}[H]
\centering
\scalebox{0.8}{
\begin{tabular}{crccccc}
  \hline
 & & Comp.1 & Comp.3 & Comp.5 & Comp.6 & Avg \\ 
  \hline
 & spec & 29.51 & 32.62 & 45.35 & 33.9 & 35.34$^{\dagger}$ \\
 \hline
 \hline
SVM  & gen & 20.23 & 25.64 & 34.01 & 46.76 & 31.66 \\ 
  \hline
 & gen & 25.75 & 21.54 & 29.75 & 31.99 & 27.26 \\ 
 RF & ens & 28.68 & 31.62 &  & 30.69 & 30.33 \\ 
&  coef. & (0.4,1) & (0.8,1) &  & (0.4,1) &  \\ 
\hline
&  gen & 27.58 & 23.26 & 27.58 & 29.48 & 26.98 \\ 
 XGB & ens & 28.85 & 28.34 &  & 31.82 & 29.67 \\ 
 & coef. & (0.1,1) & (0.3,1) &  & (0.5,1) &  \\ 
 \hline
 & \#lev. spec & 4 & 4 & 3 & 3 & 3.5 \\ 
 & \#lev. gen & 5 & 5 & 5 & 5 & 5 \\ 
 \hline
\end{tabular}
}
\caption{Severity, construction. $^{\dagger}$: best model on average. }
\end{table}

\begin{table}[H]
\centering
\scalebox{0.8}{
\begin{tabular}{crccccc}
  \hline
& & Comp.4 & Comp.6 & Comp.7 & Comp.9 & Avg \\ 
  \hline
& spec & 29.48 & 45.66 & 57.67 & 30.34 & 40.79 \\ 
\hline
\hline
 SVM & gen & 20.93 & 42.56 & 46.67 & 31.9 & 35.52 \\ 
  \hline
  & gen & 27.46 & 38.61 & 39.97 & 27.42 & 33.37 \\ 
  RF & ens & 28.73 &  & 53.62 &  & \textbf{41.17}$^{\dagger}$ \\ 
  & coef. & (1,0.9) &  & (0.5,1) &  &  \\ 
  \hline
    & gen & 27.39 & 53.02 & 39.24 & 25.95 & 36.4 \\ 
  XGB & ens & 28.74 &  & 51.27 &  & 40.01$^{\star}$ \\ 
  & coef. & (0.8,1) &  & (0.2,1) &  &  \\ 
  \hline
 & \#lev. spec & 4 & 2 & 2 & 4 & 3 \\ 
 & \#lev. gen & 5 & 5 & 5 & 5 & 5 \\ 
   \hline
\end{tabular}
}
\caption{Severity, electric T\&D. $^{\dagger}$: best model on average. \textbf{Bold}/$^{\star}$: better/within 2pts of the company-specific model.}
\end{table}

\begin{table}[H]
\centering
\scalebox{0.8}{
\begin{tabular}{crccccc}
  \hline
& & Comp.2 & Comp.3 & Comp.8 & Comp.7 & Avg \\ 
  \hline
& spec & 42.53 & 24.74 & 39.72 & 28.44 & 33.86$^{\dagger}$ \\ 
\hline
\hline
SVM & gen & 37.91 & 22.53 & 38.85 & 24.41 & 30.92 \\ 
  \hline
   & gen & 17.96 & 17.12 & 35.69 & 23.14 & 23.48 \\ 
 RF & ens & 27.87 & 24.05 & 39.81 & 24.2 & 28.98 \\ 
 & coef. & (0.2,1) & (0.7,1) & (0.7,1) & (0.1,1) &  \\ 
  \hline
   & gen & 16.75 & 23.25 & 35.27 & 21.72 & 24.25 \\ 
 XGB & ens & 27.89 & 25.36 & 39.61 & 23.7 & 29.14 \\ 
 & coef. & (0.2,1) & (1,0.8) & (0.3,1) & (0.1,1) &  \\ 
  \hline
 & \#lev. spec & 3 & 4 & 3 & 4 & 3.5 \\ 
 & \#lev. gen & 5 & 5 & 5 & 5 & 5 \\ 
   \hline
\end{tabular}
}
\caption{Severity, oil \& gas. $^{\dagger}$: best model on average.}
\end{table}

\subsection{Body Part}

\begin{table}[H]
\centering
\scalebox{0.8}{
\begin{tabular}{crccccc}
  \hline
& & Comp.1 & Comp.3 & Comp.5 & Comp.6 & Avg \\ 
  \hline
& spec & 34.14 & 26.48 & 32.09 & 31.39 & 31.03 \\ 
\hline
\hline
SVM & gen & 23.26 & 25.09 & 27.02 & 26.66 & 25.51 \\ 
\hline
  & gen & 34.14 & 33.04 & 34.68 & 35.78 & \textbf{34.41}$^{\dagger}$ \\ 
  RF & ens & 33.49 & 22.43 & 32.7 & 32.7 & 30.33$^{\star}$ \\ 
  & coef. & (0.4,1) & (0.1,1) & (0.7,1) & (0.6,1) &  \\ 
\hline
  & gen & 31.92 & 30.57 & 34.73 & 34.22 & \textbf{32.86} \\ 
  XGB & ens & 32.44 & 20.38 & 32.62 & 30.77 & 29.05$^{\star}$ \\ 
  & coef. & (0.1,1) & (0.2,1) & (0.2,1) & (0.5,1) & \\ 
\hline
 & \#lev. spec & 6 & 6 & 6 & 6 & 6 \\ 
 & \#lev. gen & 6 & 6 & 6 & 6 & 6 \\ 
   \hline
\end{tabular}
}
\caption{Body part, construction. $^{\dagger}$: best model on average. \textbf{Bold}/$^{\star}$: better/within 2pts of the company-specific model.}
\end{table}

\begin{table}[H]
\centering
\scalebox{0.8}{
\begin{tabular}{crccccc}
\hline
& & Comp.4 & Comp.6 & Comp.7 & Comp.9 & Avg \\ 
\hline
& spec & 29.25 & 27.7 & 46.34 & 23.86 & 31.79 \\ 
\hline
\hline
SVM & gen & 19.21 & 28.86 & 38.26 & 20.4 & 26.68 \\ 
\hline
 &  gen & 27.96 & 32 & 51.02 & 28.76 & \textbf{34.94}$^{\dagger}$ \\ 
RF &  ens & 27.94 &  & 50.75 & 23.56 & \textbf{34.08} \\ 
&  coef. & (0.4,1) &  & (0.1,1) & (0.4,1) &  \\ 
\hline
& gen & 28.24 & 31.12 & 46.44 & 29.24 & \textbf{33.76} \\ 
XGB & ens & 27.96 &  & 41.17 & 27.81 & \textbf{32.31} \\ 
& coef. & (0.2,1) &  & (0.1,1) & (0.5,1) &  \\ 
\hline
& \#lev. spec & 6 & 6 & 4 & 6 & 5.5 \\ 
& \#lev. gen & 6 & 6 & 6 & 6 & 6 \\ 
\hline
\end{tabular}
}
\caption{Body part, electric T\&D.  $^{\dagger}$: best model on average. \textbf{Bold}: better the company-specific model.}
\end{table}

\begin{table}[H]
\centering
\scalebox{0.8}{
\begin{tabular}{crcccc}
\hline
& & Comp.2 & Comp.8 & Comp.7 & Avg \\ 
\hline
& spec & 22.96 & 32.41 & 31.17 & 28.85 \\
\hline
\hline
SVM & gen & 22.66 & 26.23 & 22.06 & 23.65 \\
\hline
& gen & 26.11 & 32.34 & 32.21 & \textbf{30.22}$^{\dagger}$ \\ 
RF & ens & 20.31 & 32.88 & 26.09 & 26.43 \\ 
& coef. & (0.1,1) & (1,0.1) & (0.1,1) &  \\ 
\hline
& gen & 25.5 & 32.36 & 29.81 & \textbf{29.22} \\ 
XGB & ens & 16.26 & 32.28 & 30.56 & 26.37 \\ 
& coef. & (0.1,1) & (1,0.3) & (0.2,1) &  \\ 
\hline
& \#lev. spec & 6 & 6 & 6 & 6 \\ 
& \#lev. gen & 6 & 6 & 6 & 6 \\ 
\hline
\end{tabular}
}
\caption{Body part, oil \& gas. $^{\dagger}$: best model on average. \textbf{Bold}: better the company-specific model.}
\end{table}

\subsection{Injury Type}

\begin{table}[H]
\centering
\scalebox{0.8}{
\begin{tabular}{crccccc}
\hline
& & Comp.1 & Comp.3 & Comp.5 & Comp.6 & Avg \\ 
\hline
& spec & 54 & 37.7 & 33.91 & 50.07 & 43.92 \\ 
\hline
\hline
SVM & gen & 34.67 & 36.66 & 34.78 & 48.81 & 38.73 \\ 
\hline
& gen & 47.84 & 33.86 & 33.11 & 45.3 & 40.03 \\ 
RF & ens & 47.6 &  & 31.98 & 45.67 & 41.75 \\ 
& coef. & (0.2,1) &  & (0.1,1) & (0.4,1) &  \\
\hline
& gen & 46.46 & 23.99 & 31.9 & 44.7 & 36.76 \\ 
XGB & ens & 56.55 &  & 35.2 & 50.46 & \textbf{47.4}$^{\dagger}$ \\ 
& coef. & (0.6,1) &  & (0.4,1) & (0.2,1) &  \\ 
\hline
& \#lev. spec & 3 & 3 & 6 & 4 & 4 \\ 
& \#lev. gen & 11 & 11 & 11 & 11 & 11 \\ 
\hline
\end{tabular}
}
\caption{Injury type, construction. $^{\dagger}$: best model on average. \textbf{Bold}:  better the company-specific model.}
\end{table}

\begin{table}[H]
\centering
\scalebox{0.8}{
\begin{tabular}{crccccc}
\hline
& & Comp.4 & Comp.6 & Comp.7 & Comp.9 & Avg \\ 
\hline
& spec & 39.21 & 43.4 & 47.28 & 44.98 & 43.72 \\ 
\hline
\hline
SVM & gen & 42.27 & 54.59 & 60.78 & 57.14 & \textbf{53.7}$^{\dagger}$ \\ 
\hline
& gen & 26.44 & 42.41 & 56.74 & 44.16 & 42.44$^{\star}$ \\ 
RF & ens & 39.33 &  & 59.52 &  & \textbf{49.42} \\ 
& coef. & (1,0.5) &  & (1,0.2) &  &  \\ 
\hline
& gen & 28.57 & 41.6 & 51.45 & 42.49 & 41.03 \\ 
XGB & ens & 40.31 &  & 62.58 &  & \textbf{51.44} \\ 
& coef. & (1,0.8) &  & (1,0.1) &  &  \\ 
\hline
& \#lev. spec & 5 & 6 & 4 & 6 & 5.25 \\ 
& \#lev. gen & 11 & 11 & 11 & 11 & 11 \\ 
\hline
\end{tabular}
}
\caption{Injury type, electric T\&D. $^{\dagger}$: best model on average. \textbf{Bold}/$^{\star}$: better/within 2pts of the company-specific model.}
\end{table}

\begin{table}[H]
\centering
\scalebox{0.8}{
\begin{tabular}{crcccc}
\hline
& & Comp.2 & Comp.8 & Comp.7 & Avg \\ 
\hline
& spec & 35.39 & 34.04 & 40.72 & 36.72 \\ 
\hline
\hline
SVM & gen & 27.97 & 30.67 & 46.69 & 35.11$^{\star}$ \\ 
\hline
& gen & 23.72 & 32.22 & 40.52 & 32.15 \\ 
RF & ens &  & 36.82 & 40.48 & \textbf{38.65}$^{\dagger}$ \\ 
& coef. &  & (0.5,1) & (0.7,1) &  \\ 
\hline
&  gen & 23.69 & 31.01 & 39.28 & 31.33 \\ 
XGB &  ens &  & 35.09 & 41 & \textbf{38.05} \\ 
& coef. &  & (1,0.7) & (1,0.6) &  \\ 
\hline
& \#lev. spec & 3 & 10 & 8 & 7 \\ 
& \#lev. gen & 11 & 11 & 11 & 11 \\ 
\hline
\end{tabular}
}
\caption{Injury type, oil \& gas. $^{\dagger}$: best model on average. \textbf{Bold}/$^{\star}$: better/within 2pts of the company-specific model.}
\end{table}

\subsection{Accident Type}

\begin{table}[H]
\centering
\scalebox{0.8}{
\begin{tabular}{crccc}
\hline
& & Comp.3 & Comp.5 & Avg \\ 
\hline
& spec & 68.63 & 41.34 & 54.98$^{\dagger}$ \\ 
\hline
\hline
SVM & gen & 41.87 & 42.91 & 42.39 \\ 
\hline
& gen & 40.44 & 44.48 & 42.46 \\ 
RF & ens &  & 44.35 & 44.35 \\ 
& coef. &  & (1,0.7) &  \\
\hline
 & gen & 54.04 & 42.51 & 48.27 \\ 
XGB & ens &  & 42.02 & 42.02 \\ 
& coef. &  & (1,1) &  \\
 \hline
& \#lev. spec & 2 & 5 & 3.5 \\ 
& \#lev. gen & 5 & 5 & 5 \\ 
\hline
\end{tabular}
}
\caption{Accident type, construction. $^{\dagger}$: best model on average.}
\end{table}

\begin{table}[H]
\centering
\scalebox{0.8}{
\begin{tabular}{crccc}
\hline
& & Comp.4 & Comp.9 & Avg \\ 
\hline
& spec & 43.15 & 53.2 & 48.17 \\ 
\hline
\hline
SVM & gen & 36.46 & 52.71 & 44.58 \\ 
\hline
 & gen & 40.05 & 57.11 & \textbf{48.58}$^{\dagger}$ \\ 
RF &  ens & 41.29 &  & 41.29 \\ 
& coef. & (0.4,1) &  &  \\ 
\hline
 & gen & 38.13 & 57.46 & 47.8$^{\star}$ \\ 
XGB & ens & 41.08 &  & 41.08 \\ 
&  coef. & (0.4,1) &  &  \\ 
 \hline
& \#lev. spec & 5 & 4 & 4.5 \\ 
& \#lev. gen & 5 & 5 & 5 \\ 
\hline
\end{tabular}
}
\caption{Accident type, electric T\&D. $^{\dagger}$: best model on average. \textbf{Bold}/$^{\star}$: better/within 2pts of the company-specific model.}
\end{table}

\begin{table}[H]
\centering
\scalebox{0.8}{
\begin{tabular}{crcccc}
\hline
& & Comp.3 & Comp.8 & Comp.7 & Avg \\ 
\hline
& spec & 80.91 & 85 & 53.58 & 73.16$^{\dagger}$ \\ 
\hline
\hline
SVM & gen & 58.06 & 78.09 & 45.92 & 60.69 \\ 
\hline
& gen & 61.67 & 78.03 & 49.71 & 63.14 \\ 
RF & ens & 78.46 &  & 54.35 & 66.4 \\ 
& coef. & (0.1,1) &  & (1,0.1) &  \\ 
\hline
& gen & 46.65 & 76.93 & 52.16 & 58.58 \\ 
XGB & ens & 73.8 &  & 55.31 & 64.56 \\ 
& coef. & (1,0.7) &  & (1,0.7) &  \\ 
\hline
& \#lev. spec & 2 & 2 & 4 & 2.67 \\ 
& \#lev. gen & 5 & 5 & 5 & 5 \\ 
\hline
\end{tabular}
}
\caption{Accident type, oil \& gas. $^{\dagger}$: best model on average.}
\end{table}

\subsection{Energy Source}

\begin{table}[H]
\centering
\scalebox{0.8}{
\begin{tabular}{crccccc}
  \hline
& & Comp.1 & Comp.3 & Comp.5 & Comp.6 & Avg \\ 
  \hline
& spec & 71.69 & 70.97 & 68.07 & 67.82 & 69.64 \\ 
 \hline
 \hline
SVM & gen & 74.76 & 78.16 & 70.86 & 72.69 & \textbf{74.12}$^{\dagger}$ \\ 
 \hline
   & gen & 70.36 & 76.03 & 70.14 & 74.31 & \textbf{72.71} \\ 
  RF & ens & 71.05 &  & 68.02 & 68.1 & 69.06$^{\star}$ \\ 
  & coef. & (0.9,1) &  & (0.2,1) & (0.4,1) & \\ 
  \hline
    & gen & 74.33 & 83.44 & 64.62 & 72.7 & \textbf{73.77} \\ 
  XGB & ens & 71.88 &  & 66.81 & 68.47 & 69.05$^{\star}$ \\ 
  & coef. & (0.4,1) &  & (0.1,1) & (0.4,1) &  \\ 
   \hline
 & \#lev. spec & 2 & 2 & 3 & 2 & 2.25 \\ 
 & \#lev. gen & 5 & 5 & 5 & 5 & 5 \\ 
   \hline
\end{tabular}
}
\caption{Energy source, construction. $^{\dagger}$: best model on average. \textbf{Bold}/$^{\star}$: better/within 2pts of the company-specific model.}
\end{table}

\begin{table}[H]
\centering
\scalebox{0.8}{
\begin{tabular}{crcccc}
\hline
& & Comp.4 & Comp.6 & Comp.9 & Avg \\ 
\hline
& spec & 79.5 & 73.22 & 81.05 & 77.92 \\ 
\hline
\hline
SVM & gen & 76.59 & 70.61 & 85.73 & 77.64$^{\star}$ \\ 
\hline
& gen & 74.99 & 73.06 & 83.32 & 77.12$^{\star}$ \\ 
RF & ens & 77.85 & 73.81 &  & 75.83 \\ 
& coef. & (0.9,1) & (0.2,1) &  &  \\
\hline
& gen & 76.43 & 72.85 & 87.21 & \textbf{78.83}$^{\dagger}$ \\ 
XGB & ens & 79.41 & 73.52 &  & 76.47$^{\star}$ \\ 
& coef. & (0.2,1) & (0.3,1) &  &  \\ 
\hline
& \#lev. spec & 3 & 2 & 3 & 2.67 \\ 
&  \#lev. gen & 5 & 5 & 5 & 5 \\ 
\hline
\end{tabular}
}
\caption{Energy source, electric T\&D. $^{\dagger}$: best model on average. \textbf{Bold}/$^{\star}$: better/within 2pts of the company-specific model.}
\end{table}

\begin{table}[H]
\centering
\scalebox{0.8}{
\begin{tabular}{crccc}
\hline
&  & Comp.8 & Comp.7 & Avg \\ 
\hline
& spec & 68.98 & 71.8 & 70.39 \\ 
\hline
\hline
SVM & gen & 68.73 & 70.36 & 69.54$^{\star}$ \\ 
\hline
& gen & 70.43 & 71.81 & \textbf{71.12} \\ 
RF & ens & 68.27 & 72.89 & \textbf{70.58} \\ 
& coef. & (0.4,1) & (1,0.2) &  \\ 
\hline
& gen & 70.44 & 74 & \textbf{72.22}$^{\dagger}$ \\ 
XGB & ens & 68.72 & 73.25 & \textbf{70.98} \\ 
& coef. & (0.1,1) & (0.3,1) &  \\ 
\hline
& \#lev. spec & 4 & 2 & 3 \\ 
& \#lev. gen & 5 & 5 & 5 \\ 
\hline
\end{tabular}
}
\caption{Energy source, oil \& gas. $^{\dagger}$: best model on average. \textbf{Bold}/$^{\star}$: better/within 2pts of the company-specific model.}
\end{table}

\section{Per-Company Results for the Per-Domain Generic Models}\label{sec:generic_details}

Note: the ensemble (`ens') rows are left blank whenever the specific model is a SVM, as we could not use ensembling in this case (the forecast of the SVM is not probabilistic).

\subsection{Severity}

\begin{table}[H]
\centering
\scalebox{0.8}{
\begin{tabular}{crccccc}
\hline
& & Comp.5 & Comp.3 & Comp.6 & Comp.1 & Avg \\ 
\hline
& spec & 45.35 & 32.62 & 33.9 & 29.51 & 35.34$^{\dagger}$ \\ 
\hline
\hline
SVM & gen & 39.86 & 26.03 & 32.61 & 22.66 & 30.29 \\
\hline
 & gen & 34.1 & 27.7 & 32.62 & 29.74 & 31.04 \\ 
RF & ens &  & 31.2 & 33.5 & 30.77 & 31.82 \\ 
& Coeffs &  & (0.8,1) & (0.6,1) & (1,0.3) &  \\ 
\hline
 & gen & 30.84 & 26.84 & 28.33 & 28.95 & 28.74 \\ 
XGB & ens &  & 31.3 & 34.14 & 30 & 31.81 \\ 
& Coeffs &  & (0.3,1) & (1,0.6) & (0.5,1) &  \\
\hline
& \#categories spec & 3 & 4 & 3 & 4 & 3.5 \\ 
& \#categories gen & 5 & 5 & 5 & 5 & 5 \\ 
\hline
\end{tabular}
}
\caption{Severity, construction. $^{\dagger}$: best model on average.}
\end{table}

\begin{table}[H]
\centering
\scalebox{0.8}{
\begin{tabular}{crccccc}
\hline
& & Comp.7 & Comp.4 & Comp.9 & Comp.6 & Avg \\ 
\hline
& spec & 57.67 & 29.48 & 30.34 & 45.66 & 40.79 \\ 
\hline
\hline
SVM & gen & 36 & 30.47 & 24.62 & 52.06 & 35.79 \\ 
\hline
 & gen & 47.19 & 30.91 & 28.5 & 58.53 & \textbf{41.28} \\ 
RF & ens & 54.8 & 32.97 &  &  & \textbf{43.88}$^{\dagger}$ \\ 
& Coeffs & (1,0.6) & (1,0.9) &  &  &  \\ 
\hline
 & gen & 44.23 & 30.59 & 26.49 & 57.44 & 39.69$^{\star}$ \\ 
XGB & ens & 54.95 & 26.96 &  &  & \textbf{40.95} \\ 
& Coeffs & (0.4,1) & (0.1,1) &  &  &  \\ 
\hline
& \#categories spec & 2 & 4 & 4 & 2 & 3 \\ 
& \#categories gen & 5 & 5 & 5 & 5 & 5 \\ 
\hline
\end{tabular}
}
\caption{Severity, elec. $^{\dagger}$: best model on average. \textbf{Bold}/$^{\star}$: better/within 2pts of the company-specific model.}
\end{table}

\begin{table}[H]
\centering
\scalebox{0.8}{
\begin{tabular}{crccccc}
\hline
& & Comp.7 & Comp.3 & Comp.8 & Comp.2 & Avg \\ 
\hline
& spec & 28.44 & 24.74 & 39.72 & 42.53 & 33.86$^{\dagger}$ \\ 
\hline
\hline
SVM & gen & 26.7 & 21.05 & 40.83 & 27.31 & 28.97 \\ 
\hline
 & gen & 26.22 & 19.82 & 35.7 & 22.59 & 26.08 \\ 
RF &  ens & 27.59 & 22.38 & 40.22 & 32.86 & 30.76 \\ 
&  Coeffs & (0.8,1) & (1,0.9) & (1,0.8) & (0.7,1) &  \\ 
\hline
 & gen & 23.82 & 19.7 & 33.09 & 20.15 & 24.19 \\ 
XGB & ens & 27.97 & 25.73 & 40.1 & 31.96 & 31.44 \\ 
& Coeffs & (0.4,1) & (1,0.7) & (1,0.2) & (0.7,1) &  \\ 
\hline
& \#categories spec & 4 & 4 & 3 & 3 & 3.5 \\ 
& \#categories gen & 5 & 5 & 5 & 5 & 5 \\ 
\hline
\end{tabular}
}
\caption{Severity, oil \& gas. $^{\dagger}$: best model on average.}
\end{table}

\subsection{Body part}

\begin{table}[H]
\centering
\scalebox{0.8}{
\begin{tabular}{crccccc}
\hline
& & Comp.5 & Comp.3 & Comp.6 & Comp.1 & Avg \\ 
\hline
& spec & 32.09 & 26.48 & 31.39 & 34.14 & 31.03 \\ 
\hline
\hline
SVM & gen & 31.08 & 28.14 & 31.92 & 34.13 & \textbf{31.32} \\ 
\hline
 & gen & 32.19 & 27.06 & 33.64 & 34.34 & \textbf{31.81} \\ 
RF & ens & 31.23 & 25.77 & 35.14 & 29.41 & 30.39$^{\star}$ \\
&  Coeffs & (0.1,1) & (0.2,1) & (0.6,1) & (0.1,1) &  \\ 
\hline
 & gen & 33.6 & 29.91 & 32.48 & 32.92 & \textbf{32.23}$^{\dagger}$ \\ 
XGB & ens & 32.34 & 20.72 & 32.33 & 31.41 & 29.2$^{\star}$ \\ 
& Coeffs & (0.1,1) & (0.2,1) & (0.5,1) & (0.1,1) &  \\ 
\hline
& \#categories spec & 6 & 6 & 6 & 6 & 6 \\ 
& \#categories gen & 6 & 6 & 6 & 6 & 6 \\ 
\hline
\end{tabular}
}
\caption{Body part, construction. $^{\dagger}$: best model on average. \textbf{Bold}/$^{\star}$: better/within 2pts of the company-specific model.}
\end{table}

\begin{table}[H]
\centering
\scalebox{0.8}{
\begin{tabular}{crccccc}
\hline
& & Comp.7 & Comp.4 & Comp.9 & Comp.6 & Avg \\ 
\hline
& spec & 46.34 & 29.25 & 23.86 & 27.7 & 31.79 \\ 
\hline
\hline
SVM &  gen & 34.02 & 18.16 & 19.6 & 29.17 & 25.24 \\ 
\hline
 & gen & 48.21 & 26.31 & 25.71 & 31.97 & \textbf{33.05} \\ 
RF & ens & 39.52 & 28.63 & 19.59 &  & 29.25 \\ 
& Coeffs & (0.1,1) & (0.1,1) & (0.1,1) &  &  \\ 
\hline
 & gen & 53.03 & 28.55 & 26.56 & 32.7 & \textbf{35.21}$^{\dagger}$ \\ 
XGB & ens & 49.41 & 29.72 & 25.01 &  & \textbf{34.71} \\ 
& Coeffs & (1,1) & (0.6,1) & (0.2,1) &  &  \\
\hline
& \#categories spec & 4 & 6 & 6 & 6 & 5.5 \\ 
& \#categories gen & 6 & 6 & 6 & 6 & 6 \\  
\hline
\end{tabular}
}
\caption{Body part, elec. $^{\dagger}$: best model on average. \textbf{Bold}/$^{\star}$: better/within 2pts of the company-specific model.}
\end{table}

\begin{table}[H]
\centering
\scalebox{0.8}{
\begin{tabular}{crcccc}
\hline
& & Comp.7 & Comp.8 & Comp.2 & Avg \\ 
\hline
& spec & 31.17 & 32.41 & 22.96 & 28.85$^{\dagger}$ \\ 
\hline
\hline
SVM & gen & 27.22 & 25.91 & 20.84 & 24.66 \\ 
\hline
 & gen & 29.64 & 31.8 & 25.12 & 28.85$^{\dagger}$ \\ 
RF & ens & 30.15 & 31.66 & 20.81 & 27.54$^{\star}$ \\ 
& Coeffs & (0.1,1) & (0.1,1) & (0.4,1) &  \\ 
\hline
 & gen & 29.69 & 32.36 & 23.72 & 28.59$^{\star}$ \\ 
XGB & ens & 31.84 & 32.1 & 19.28 & 27.74$^{\star}$ \\ 
& Coeffs & (1,0.5) & (1,0.1) & (0.2,1) &  \\ 
\hline
& \#categories spec & 6 & 6 & 6 & 6 \\ 
& \#categories gen & 6 & 6 & 6 & 6 \\ 
\hline
\end{tabular}
}
\caption{Body part, oil \& gas. $^{\dagger}$: best model on average. \textbf{Bold}/$^{\star}$: better/within 2pts of the company-specific model.}
\end{table}

\subsection{Injury type}

\begin{table}[H]
\centering
\scalebox{0.8}{
\begin{tabular}{crccccc}
\hline
& & Comp.5 & Comp.3 & Comp.6 & Comp.1 & Avg \\ 
\hline
& spec & 33.91 & 37.7 & 50.07 & 54 & 43.92 \\ 
\hline
\hline
SVM & gen & 34.16 & 36.31 & 51.7 & 48.97 & 42.78$^{\star}$ \\ 
\hline
 & gen & 33.56 & 33.91 & 49.34 & 51.64 & 42.11$^{\star}$ \\ 
RF & ens & 33.38 &  & 48.57 & 54.42 & \textbf{45.46}$^{\dagger}$ \\ 
& Coeffs & (0.1,1) &  & (0.1,1) & (1,0.2) &  \\
\hline
 & gen & 33.3 & 38.19 & 48.17 & 50.54 & 42.55$^{\star}$ \\ 
XGB & ens & 34.74 &  & 47.08 & 54.53 & \textbf{45.45} \\ 
& Coeffs & (0.3,1) &  & (0.1,1) & (0.5,1) &  \\ 
\hline
& \#categories spec & 6 & 3 & 4 & 3 & 4 \\ 
& \#categories gen & 6 & 6 & 6 & 6 & 6 \\ 
\hline
\end{tabular}
}
\caption{Injury type, construction. $^{\dagger}$: best model on average. \textbf{Bold}/$^{\star}$: better/within 2pts of the company-specific model.}
\end{table}

\begin{table}[H]
\centering
\scalebox{0.8}{
\begin{tabular}{crccccc}
\hline
&& Comp.7 & Comp.4 & Comp.9 & Comp.6 & Avg \\ 
\hline
& spec & 47.28 & 39.21 & 44.98 & 43.4 & 43.72 \\
\hline
\hline
SVM &  gen & 57.28 & 28.12 & 44.54 & 41.7 & 42.91$^{\star}$ \\ 
\hline
 & gen & 53.99 & 29.2 & 43.07 & 40.47 & 41.68$^{\star}$ \\ 
RF & ens & 51.33 & 38.87 &  &  & \textbf{45.1} \\ 
& Coeffs & (0.8,1) & (0.4,1) &  &  &  \\ 
\hline
 & gen & 49.62 & 29.63 & 40.26 & 46.42 & 41.48 \\ 
XGB & ens & 59.48 & 39.09 &  &  & \textbf{49.28}$^{\dagger}$ \\ 
& Coeffs & (1,0.3) & (1,0.3) &  &  &  \\ 
\hline
& \#categories spec & 4 & 5 & 6 & 6 & 5.25 \\ 
& \#categories gen & 8 & 8 & 8 & 8 & 8 \\ 
\hline
\end{tabular}
}
\caption{Injury type, elec. $^{\dagger}$: best model on average. \textbf{Bold}/$^{\star}$: better/within 2pts of the company-specific model.}
\end{table}

\begin{table}[H]
\centering
\scalebox{0.8}{
\begin{tabular}{crcccc}
\hline
& & Comp.7 & Comp.8 & Comp.2 & Avg \\ 
\hline
& spec & 40.72 & 34.04 & 35.39 & 36.72 \\ 
\hline
\hline
SVM & gen & 50.26 & 33.24 & 18.18 & 33.89 \\ 
\hline
 & gen & 39.57 & 33.87 & 25.02 & 32.82 \\ 
RF & ens & 41.69 & 36.25 &  & \textbf{38.97} \\ 
& Coeffs & (1,0.7) & (0.8,1) &  &  \\ 
\hline
 & gen & 38.32 & 32.64 & 26.07 & 32.34 \\ 
XGB & ens & 42.88 & 36.7 &  & \textbf{39.79}$^{\dagger}$ \\ 
& Coeffs & (1,0.7) & (1,0.1) &  &  \\ 
\hline
& \#categories spec & 8 & 10 & 3 & 7 \\ 
& \#categories gen & 11 & 11 & 11 & 11 \\ 
\hline
\end{tabular}
}
\caption{Injury type, oil \& gas. $^{\dagger}$: best model on average. \textbf{Bold}/$^{\star}$: better/within 2pts of the company-specific model.}
\end{table}

\subsection{Accident type}

\begin{table}[H]
\centering
\scalebox{0.8}{
\begin{tabular}{crccc}
\hline
& & Comp.5 & Comp.3 & Avg \\ 
\hline
& spec & 41.34 & 68.63 & 54.98$^{\dagger}$ \\ 
\hline
\hline
SVM & gen & 44.25 & 40.15 & 42.2 \\ 
\hline
 & gen & 41.37 & 43.48 & 42.42 \\ 
RF & ens & 40.8 &  & 40.8 \\ 
& Coeffs & (0.1,1) &  &  \\ 
\hline
 & gen & 43.21 & 55.21 & 49.21 \\ 
XGB & ens & 43.4 &  & 43.4 \\ 
& Coeffs & (1,0.1) &  &  \\ 
\hline
& \#categories spec & 5 & 2 & 3.5 \\ 
& \#categories gen & 5 & 5 & 5 \\ 
\hline
\end{tabular}
}
\caption{Accident type, construction. $^{\dagger}$: best model on average.}
\end{table}

\begin{table}[H]
\centering
\scalebox{0.8}{
\begin{tabular}{crccc}
\hline
& & Comp.4 & Comp.9 & Avg \\ 
\hline
& spec & 43.15 & 53.2 & 48.17 \\ 
\hline
\hline
SVM & gen & 39.45 & 50.22 & 44.84 \\ 
\hline
 & gen & 44.13 & 56.28 & \textbf{50.2}$^{\dagger}$ \\ 
RF & ens & 39.72 &  & 39.72 \\ 
& Coeffs & (0.3,1) &  &  \\ 
\hline
 & gen & 40.96 & 58.21 & \textbf{49.58} \\ 
XGB & ens & 43.53 &  & 43.53 \\ 
& Coeffs & (0.4,1) &  &  \\ 
\hline
& \#categories spec & 5 & 4 & 4.5 \\ 
&  \#categories gen & 5 & 5 & 5 \\ 
\hline
\end{tabular}
}
\caption{Accident type, elec. $^{\dagger}$: best model on average. \textbf{Bold}/$^{\star}$: better/within 2pts of the company-specific model.}
\end{table}

\begin{table}[H]
\centering
\scalebox{0.8}{
\begin{tabular}{crcccc}
\hline
& & Comp.7 & Comp.3 & Comp.8 & Avg \\ 
\hline
& spec & 53.58 & 80.91 & 85 & 73.16$^{\dagger}$ \\ 
\hline
\hline
SVM &  gen & 55.04 & 59.76 & 79.75 & 64.85 \\ 
\hline 
 & gen & 51.77 & 58.06 & 82.53 & 64.12 \\ 
RF & ens & 53.02 & 78.46 &  & 65.74 \\ 
& Coeffs & (1,0.1) & (0.1,1) &  &  \\
\hline
 & gen & 49.03 & 62.16 & 77.93 & 63.04 \\ 
XGB & ens & 55.7 & 78.56 &  & 67.13 \\ 
& Coeffs & (1,0.9) & (0.1,1) &  &  \\ 
\hline
& \#categories spec & 4 & 2 & 2 & 2.67 \\ 
& \#categories gen & 4 & 4 & 4 & 4 \\ 
\hline
\end{tabular}
}
\caption{Accident type, oil \& gas. $^{\dagger}$: best model on average.}
\end{table}

\subsection{Energy source}

\begin{table}[H]
\centering
\scalebox{0.8}{
\begin{tabular}{crccccc}
\hline
& & Comp.5 & Comp.3 & Comp.6 & Comp.1 & Avg \\ 
\hline
& spec & 68.07 & 70.97 & 67.82 & 71.69 & 69.64 \\ 
\hline
\hline
SVM & gen & 70.3 & 76.99 & 73.32 & 74.5 & \textbf{73.78} \\ 
\hline
 & gen & 68.17 & 79.98 & 74.28 & 73.21 & \textbf{73.91}$^{\dagger}$ \\ 
RF & ens & 67.85 &  & 69.62 & 71.45 & 69.64$^{\star}$ \\ 
& Coeffs & (0.1,1) &  & (0.9,1) & (0.4,1) &  \\ 
\hline
 & gen & 62.88 & 73.28 & 74.91 & 73.09 & \textbf{71.04} \\ 
XGB & ens & 68.05 &  & 70.75 & 71.9 & \textbf{70.23} \\ 
& Coeffs & (0.1,1) &  & (0.7,1) & (0.5,1) &  \\ 
\hline
& \#categories spec & 3 & 2 & 2 & 2 & 2.25 \\ 
& \#categories gen & 3 & 3 & 3 & 3 & 3 \\ 
\hline
\end{tabular}
}
\caption{Energy source, construction. $^{\dagger}$: best model on average. \textbf{Bold}/$^{\star}$: better/within 2pts of the company-specific model.}
\end{table}

\begin{table}[H]
\centering
\scalebox{0.8}{
\begin{tabular}{crcccc}
\hline
& & Comp.4 & Comp.9 & Comp.6 & Avg \\ 
\hline
& spec & 79.5 & 81.05 & 73.22 & 77.92 \\ 
\hline
\hline
SVM & gen & 78.31 & 85.83 & 72.46 & \textbf{78.87}$^{\dagger}$ \\ 
\hline
 & gen & 80.13 & 83.73 & 71.96 & \textbf{78.61} \\ 
RF & ens & 79.75 &  & 73.22 & 76.48$^{\star}$ \\ 
& Coeffs & (0.5,1) &  & (0.1,1) &  \\ 
\hline
 & gen & 75.63 & 82.34 & 68.86 & 75.61 \\ 
XGB & ens & 77.15 &  & 72.91 & 75.03 \\ 
& Coeffs & (1,0.8) &  & (0.1,1) &  \\ 
\hline
& \#categories spec & 3 & 3 & 2 & 2.67 \\ 
& \#categories gen & 3 & 3 & 3 & 3 \\ 
\hline
\end{tabular}
}
\caption{Energy source, elec. $^{\dagger}$: best model on average. \textbf{Bold}/$^{\star}$: better/within 2pts of the company-specific model.}
\end{table}

\begin{table}[H]
\centering
\scalebox{0.8}{
\begin{tabular}{crccc}
\hline
& & Comp.7 & Comp.8 & Avg \\ 
\hline
& spec & 71.8 & 68.98 & 70.39 \\ 
\hline
\hline
SVM & gen & 49.94 & 61.33 & 55.63 \\ 
\hline
 & gen & 69.34 & 72.11 & \textbf{70.72}$^{\dagger}$ \\ 
RF & ens & 70.44 & 67.8 & 69.12$^{\star}$ \\ 
& Coeffs & (1,0.5) & (0.4,1) &  \\ 
\hline
 & gen & 72.58 & 68.12 & 70.35$^{\star}$ \\ 
XGB & ens & 72.06 & 68.69 & 70.38$^{\star}$ \\ 
& Coeffs & (0.2,1) & (0.1,1) &  \\ 
\hline
& \#categories spec & 2 & 4 & 3 \\ 
& \#categories gen & 4 & 4 & 4 \\ 
\hline
\end{tabular}
}
\caption{Energy source, oil \& gas. $^{\dagger}$: best model on average. \textbf{Bold}/$^{\star}$: better/within 2pts of the company-specific model.}
\end{table}

\end{document}